\documentclass[10pt]{article}
\usepackage[legalpaper, margin=1in]{geometry}
\usepackage{amsmath,amsfonts}
\usepackage{algorithmic}
\usepackage{algorithm}
\usepackage{array}
\usepackage[caption=false,font=normalsize,labelfont=sf,textfont=sf]{subfig}
\usepackage{textcomp}
\usepackage{stfloats}
\usepackage{verbatim}
\usepackage{graphicx}
\usepackage{cite}
\usepackage{xcolor}
\usepackage{multirow}
\usepackage{wrapfig}

\newcommand{\alg}{Q-TART}
\usepackage{booktabs}
\usepackage{enumitem}
\usepackage{subfig}
\usepackage{bbm}
\usepackage{multirow}
\usepackage[hyphens]{url}
\usepackage{hyperref}
\begin{document}


\title{Q-TART: Quickly Training for Adversarial Robustness and in-Transferability}

\author{Madan Ravi Ganesh, Salimeh Yasaei Sekeh, and Jason J. Corso}%

\date{}
\maketitle

\begin{abstract}
Raw deep neural network (DNN) performance is not enough; in real-world settings, computational load, training efficiency and adversarial security are just as or even more important.
We propose to simultaneously tackle \textbf{P}erformance, \textbf{E}fficiency, and \textbf{R}obustness, using our proposed algorithm \alg, Quickly Train for Adversarial Robustness and in-Transferability.
\alg{} follows the intuition that samples highly susceptible to noise strongly affect the decision boundaries learned by DNNs, which in turn degrades their performance and adversarial susceptibility.
By identifying and removing such samples, we demonstrate improved performance and adversarial robustness while using only a subset of the training data.
Through our experiments we highlight \alg’s high performance across multiple Dataset-DNN combinations, including ImageNet, and provide insights into the complementary behavior of \alg{} alongside existing adversarial training approaches to increase robustness by over $1.3\%$ while using up to $17.9\%$ less training time.

\end{abstract}

\section{Introduction}
The ability to learn patterns from large-scale data while not requiring explicit analytical modelling has made deep neural networks (DNN) much sought after in recent years.
The development of DNN solutions is a time and resource intensive process which often focuses solely on improving the performance of the learned model.
However, to deal with the rigors of the real-world DNNs should not only be 1) highly accurate, but 2) efficient to develop, and 3) robust to adversaries as well.
Since each of these properties fulfill unique targets they are often handled as separate issues.
Ideally, by jointly constraining the development process to satisfy Performance (P), Efficiency (E) and Robustness (R), or PER goals, we can significantly reduce the cost of developing DNN solutions, that are both highly accurate as well as secure from attacks.

To the best of our knowledge, there are no works that simultaneously address all three PER targets.
Methods that focus on efficiency, like distributed training~\cite{rajbhandari2020zero,devlin2019bert} assume the availability of large-scale hardware while low precision computations~\cite{gupta2015deep,sun2020ultra} rarely match the potential of their high precision counterparts.
From an adversarial robustness perspective, only a subset of works under adversarial training address the idea of efficiently imparting robustness~\cite{DBLP:journals/corr/abs-2103-03076,DBLP:conf/iclr/WongRK20,shafahi2019adversarial}.
Their main focus is the choice of algorithm to generate adversaries efficiently.
However, they retain the entire training dataset in memory which leads to overheads in loading, preprocessing and training time.
A common theme across these different categories of solutions is their focus on tackling at most two of the three desired PER goals.

To effectively tackle all three PER targets simultaneously, we propose \alg, a method to Quickly Train for Adversarial Robustness and in-~Transferability.
Our algorithm is built on the assumption that there exists a subset of the original training data that negatively impacts the final learned model~\cite{DBLP:journals/corr/LapedrizaPBT13}.
In \alg, we use a function of the distance between features, specifically between the original inputs and their noise-perturbed counterparts, as a heuristic to identify and remove the subset of samples that negatively impact performance.
In doing so, we use only a subset of the available data to train the model which reduces the overall training time while improving the generalization performance.

\begin{figure}[t!]
\centering
\subfloat[][Distribution of feature distances]{\includegraphics[width=0.49\columnwidth]{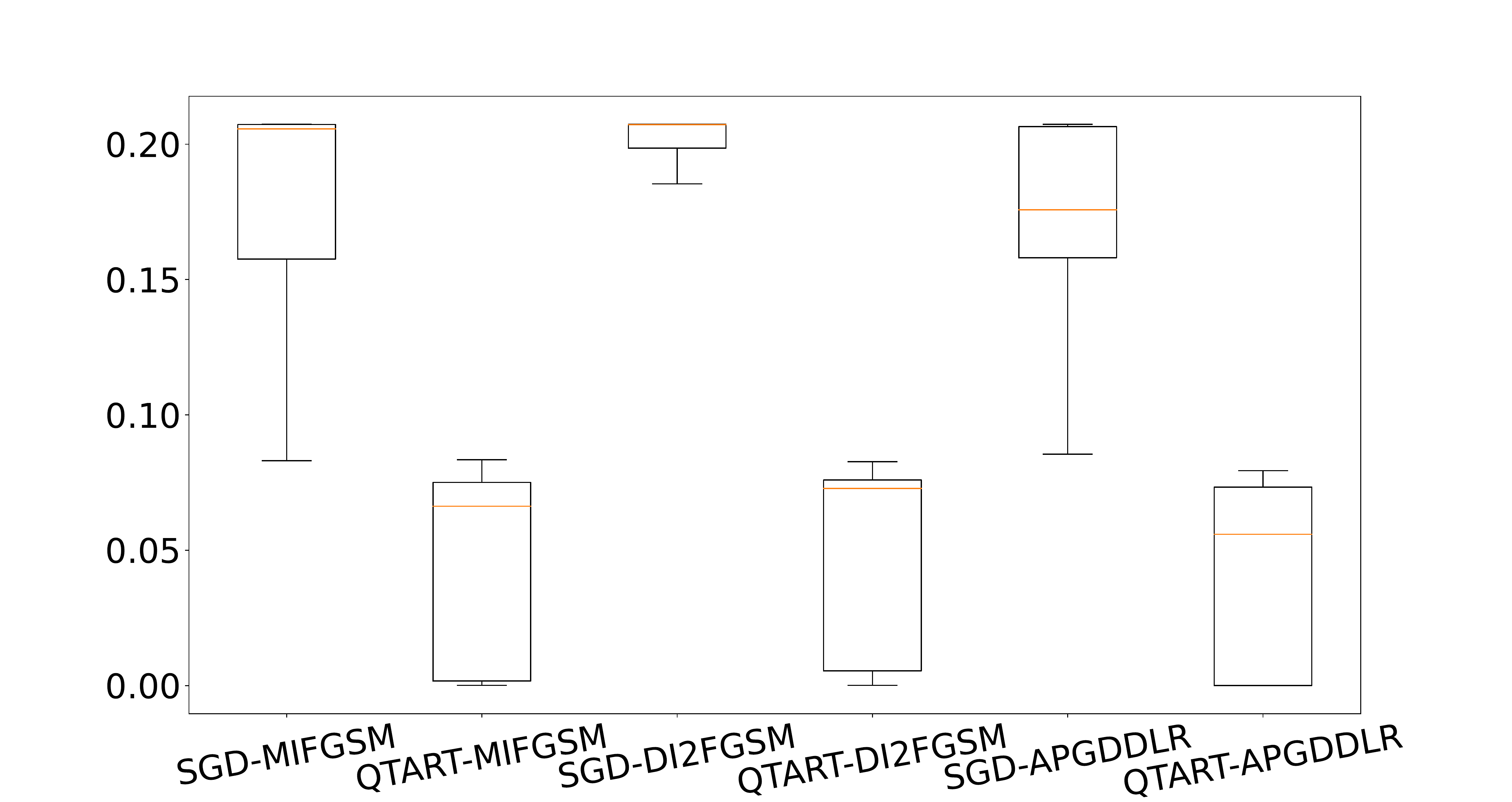}\label{fig:r50_feature_distributions}}
\subfloat[][Baseline]{\includegraphics[width=0.25\columnwidth]{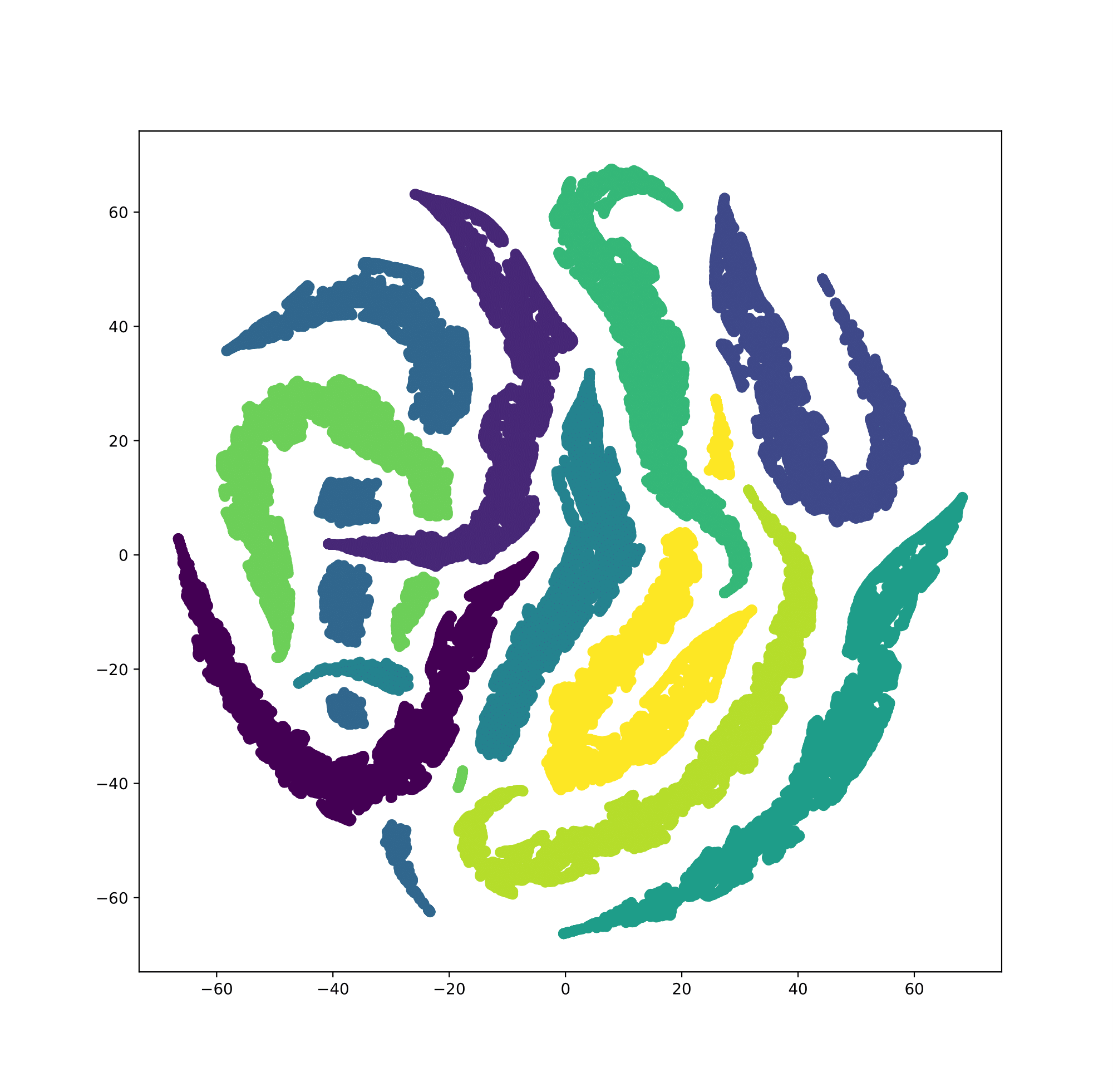}\label{fig:r50_baseline}}
\subfloat[][\alg{}]{\includegraphics[width=0.25\columnwidth]{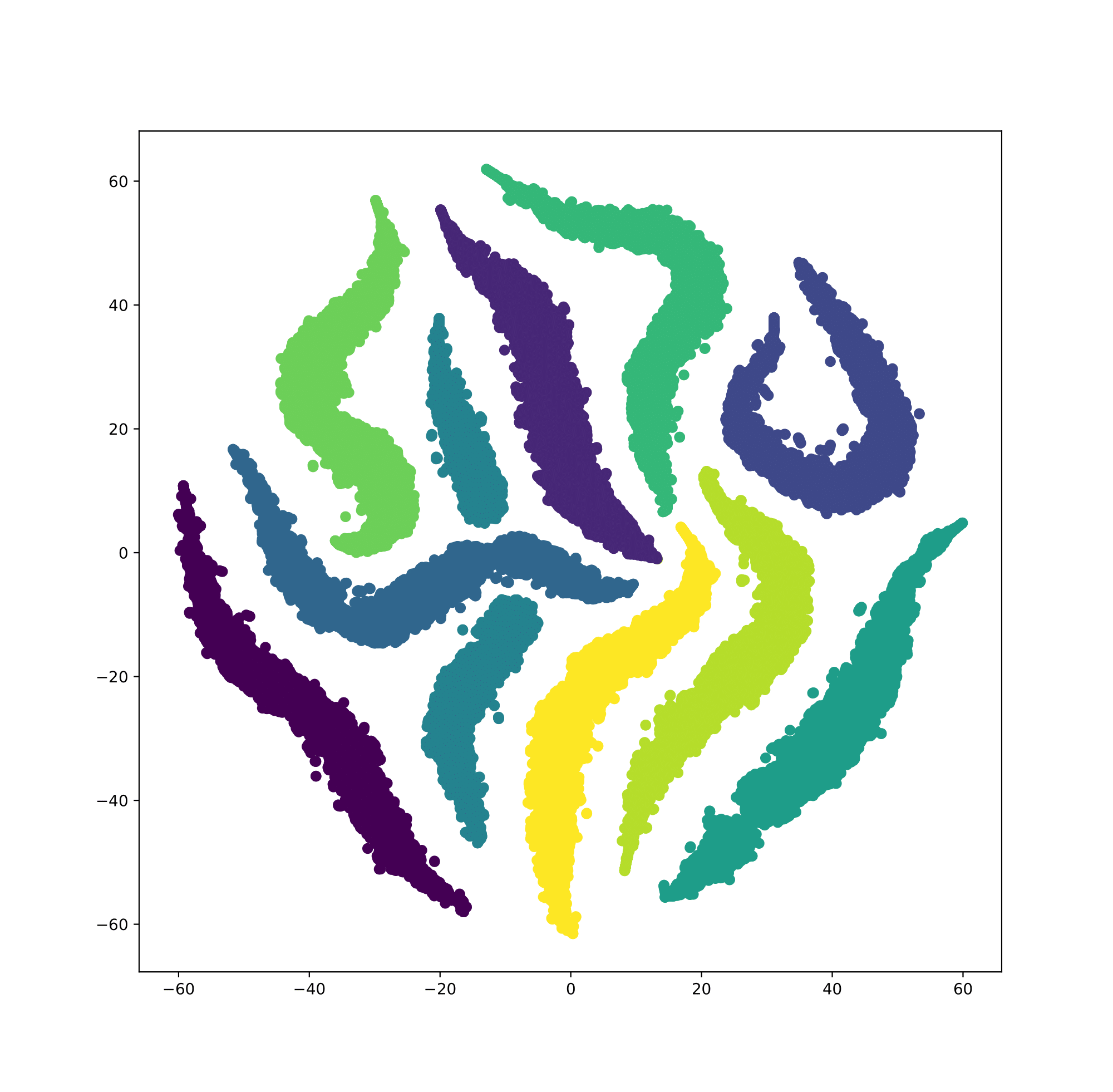}\label{fig:r50_qtart}}
\caption{Regularization effect of \alg{} on the learned features in ResNet50-CIFAR10. Fig.~\ref{fig:r50_feature_distributions} compares the distribution of feature distances, computed from the last layer, between mini-batch SGD and \alg{} in the presence of three adversarial attacks. 
We clearly observe a reduction in the range of the distribution of feature distances, a measure of robustness, when \alg{} is used to train the model. Figs.~\ref{fig:r50_baseline} and~\ref{fig:r50_qtart} illustrate the relatively smooth and contiguous tSNE clusters formed in \alg{} when compared to SGD}
\label{fig:reg_qtart}
\end{figure}

In contrast to conventional curriculum learning, Q-TART uses noise-injection and the subsequent feature distance as a measure to select the samples that are retained for actual training (those samples expected to improve adversarial robustness of the final model).
One way to conceptualize our approach is as the interaction between dataset and DNN, where the DNN is held constant while regularizing the dataset by penalizing certain samples and removing them.
Fig.~\ref{fig:reg_qtart} illustrates the expected behavior of using \alg{} to train DNNs in the presence of adversaries.
In this work, we analyse adversarial robustness from two perspectives, where the source and target are the same model and the alternative where the source model can be different from the target.
While the first scenario establishes robustness to a known model, the second scenario measures the robustness of DNNs to attacks designed on a variety of backbones, a property we define as in-Transferability.
Through our experimental validation, we highlight the difference in how robustness is imparted to a DNN using \alg{} when compared to standard adversarial training~\cite{madry2018towards,tramer2018ensemble,carlini2017towards}.
Keeping this distinction in mind, we build \alg{} atop efficient adversarial training regimes and highlight its complementary behavior.

\noindent To summarize, our contributions in this paper are,
\begin{itemize}[topsep=0pt,itemsep=0ex,partopsep=0ex,parsep=0ex]
    \item \alg, a new methodology that simultaneously targets improved performance, efficiency in the training phase, and robustness to adversarial attacks.
    \item Improved robustness across a variety of adversarial attacks, when adversaries are generated with a known source model as well as when adversaries are transferred from other models, a property we term in-Transferability.
    \item Demonstration of complementary behavior alongside efficient adversarial training regimes to boost robustness to adversarial attacks.
\end{itemize}
\section{Related Works}
\label{sec:related_works}
To the best of our knowledge, there is no prior work that simultaneously targets improvements in performance, efficiency and robustness. 
We restrict our discussion of related works to the general curriculum learning domain, since they are closest in spirit to addressing performance and efficiency targets, and works in the sub-domain of efficient adversarial training.

\subsection{General Curriculum Learning}
In its nascent stage, curriculum learning was defined as an approach to organize and present data to machine learning models to improve their learning process and performance.
A crucial point of emphasis was their fast convergence to a high quality solution~\cite{bengio2009curriculum,graves2017automated}.
Subsequent works focused on various approaches to organize and schedule data while relaxing the constraints on faster convergence~\cite{hacohen2019power,zhou2018minimax,DBLP:conf/bmvc/GaneshC20,jiang2018mentornet}.
Recently, there has been a shift towards stronger focus on using feedback from the model being trained, to modify the training regime, in parallel with reducing the amount of data used to train the model~\cite{zhou2020curriculum,zhou2021curriculum}.
However, across all works in curriculum learning there has always been an emphasis on improving the generalization performance of the final solution, with little attention given to adversarial robustness.
In our work, we consider robustness to adversarial attacks a key trait required of DNNs, especially when we consider their application in a safety critical real-world contexts.

From a methodological point of view, our approach uses additive noise to identify and remove samples that create adversarial vulnerability.
This is distinct from the direct use of gradients, loss value, predictions or the change in those values to identify difficult samples~\cite{zhou2018minimax,toneva2018empirical,smith2014instance,loshchilov2015online}.
Additionally, in \alg{} we use hard sampling to permanently remove samples from the training set instead of recycling them during the training phase~\cite{zhou2020curriculum}.
Our approach is more similar to the hard sampling performed in~\cite{DBLP:journals/corr/LapedrizaPBT13,chitta2021training}.
Furthermore, since our approach focuses on differences in the feature embedding space as the primary means to highlight samples that need to be removed, it is easily extensible to different architectures and applications.

\subsection{Adversarial Training}
Adversarial training approaches expose the DNN model to a variety of adversarial perturbations during the training phase to increase their robustness to adversarial attacks~\cite{madry2018towards,zhang2019theoretically}.
A  number of adversarial training approaches emphasize various modifications to constraints used in the algorithms that generate adversaries including gradually increasing the strength of adversaries to improve robustness~\cite{cai2018curriculum}, using the least adversarial data among confidently misclassified samples~\cite{zhang2020attacks} and others~\cite{wang2019convergence}.
However, they rarely provide time or efficiency comparisons to standard or alternative adversarial training regimes.

A more recent line of works tackle the problem of efficient adversarial training, including Wang \textit{et al.}~\cite{DBLP:journals/corr/abs-2103-03076} who propose a dynamic and efficient adversarial training methodology that automatically learns to adjust the magnitude of perturbations during the training process.
While their theoretical analysis, computational complexity, and performance comparisons offer strong insights, their results are limited to fixed DNN backbones.
Shafahi \textit{et al.}~\cite{shafahi2019adversarial} offer an inexpensive alternative of recycling gradient computations performed during backpropagation to generate adversarial examples.
Wong \textit{et al.}~\cite{DBLP:conf/iclr/WongRK20} review FGSM-based adversarial training and offer multiple key suggestions that extend FGSM's viability to quickly obtain highly robust DNNs.
Each of the above methods that propose a more efficient adversarial training approach focus on modifying the algorithm used to generate adversaries while retaining the complete training set.
However, in \alg{} we address training efficiency by directly reducing the training data available, thus offering a complementary approach that can work alongside any traditional efficient adversarial training algorithm.

\section{Q-TART}
\label{sec:qtart}


\subsection{Standard Setup}
\label{subsec:standard_setup}

When training an $L$ layer DNN for classification, the input variables are denoted by $\{(x_i,y_i)\}_{i=1}^{N} \sim (X,Y)$, where $N$ represents the total number of samples.
Here, $x_i \in \mathbb{R}^{H\times W\times 3}$, is the input RGB image and $y_i \in \{1,2,\ldots,C\}$ is the ground-truth label in a dataset with $C$ classes.
The output of layer $l$ is denoted by,
\begin{equation}
f^{(l)}(x^{(l-1)}_i) = \sigma(W^{(l)} x^{(l-1)}_i + b^{(l)})\enspace,     
\end{equation}
assuming an activation function $\sigma()$, $f^{(l)} \in \mathbb{R}^{N \times O^{(l)} \times h^{(l)} \times w^{(l)}}$,  
where $O^{(l)}$ denotes the output dimension of layer $l$, $h^{(l)}$, $w^{(l)}$, $W^{(l)}$, and $ b^{(l)}$ represent the output height, width, weights and biases of layer $l$, respectively.
The general loss function used to train this setup is,
\begin{equation}
    \label{eq:general_std_loss}
    \mathcal{L}(X,Y) = \min_W \frac{1}{N}\sum_{i=1}^N \ell(F(x_i), y_i)\enspace,
\end{equation}
where $F()$ denotes the output of the entire DNN. 
For classification, $\ell()$ becomes the multi-class cross-entropy loss.


\subsection{Key Notations}
\label{subsec:key_notations}
\begin{itemize}[topsep=0pt,itemsep=-1ex,partopsep=1ex,parsep=1ex]
    \item $\epsilon$ : Smoothing value for ground-truth variables when loss is evaluated.
    \item $\tau$ : Training epoch at which \alg{} is applied.
    \item $m_i$ : Binary value indicating whether the $i^{th}$ sample is retained or removed.
    \item $H()$ : Function used to project features to lower dimensions.
    \item $D()$ : Distance function.
    \item $\tilde{O}^{(l)}$ : Sensitivity-based subset of filters used to capture features.
    \item $\xi_i$ : Overall instability score for each sample in the training data.
    \item $\alpha()$ : Window function that assigns multipliers to instability values from different layers.
    \item $\gamma$ : Number of samples removed from training data.
\end{itemize}

\subsection{Proposed Algorithm}
\label{subsec:algorithm}
In \alg{}, we focus on removing a subset of the training data that negatively impacts performance.
We begin by training a DNN using the complete training dataset up to $\tau$ epochs. 
At the chosen epoch $\tau << E$, where $E$ is the total number of training epochs, we compare the distance between features, specifically between standard inputs and their noise-perturbed counterparts.
Here, the noise-perturbed counterparts are generated using additive gaussian noise on the input images.
A large distance between the features highlights samples highly susceptible to noise. 
We use the distance values to generate a binary mask and remove these samples from the dataset.
We posit that the dataset is regularized by the removal of noisy samples, which translates to improved robustness of the learned features as well as their overall quality, measured through performance.
An explanation of the exact processes underlying \alg{} is provided below.

\subsubsection{Setup}
In \alg{} we modify the loss function used to learn the weights of the DNN by masking the contributions from the noisy subset of data: 
\begin{equation}
    \label{eq:qtart_loss}
    \mathcal{L}(X,Y) = \min_W \frac{1}{\vert \vert m \vert \vert_{0}}\sum_{i=1}^N m_i\ell_{\epsilon}(F(x_i), y_i),~ m_i \in \{0,1\}\enspace.
\end{equation}
Here, $m \in \{0,1\}^N$ is the binary mask vector defined using our heuristic based on the distance between features.
Once we determine the value of $m$ at epoch $\tau$, it remains fixed throughout the remaining training epochs.
An extremely small value of $\tau$ would capture features that aren't coherent while large values of $\tau$ would significantly reduce the efficiency gain we expect.  
Instead, we choose a relatively small value but balanced value of $\tau \in \{50,100\}$ to obtain coherent features and maximize our gain in efficiency.
In addition, we use the cross-entropy loss modified by label smoothing~\cite{szegedy2016rethinking} ($\ell_{\epsilon}$), where the smoothing operation on the one-hot ground-truth vector can be defined as,
\begin{equation}
    y_{ls} = (1-\epsilon) \times \mathbbm{1}_{y} + \frac{\epsilon}{C}\enspace.
\end{equation}
Here $\epsilon$ is defined as the smoothing value and $\mathbbm{1}_{y}$ is a one-hot vector at the ground-truth label. 

\subsubsection{Capturing Feature Distance}
To ascertain the value of $m$, we begin by capturing the distance between features, specifically between the original input and their noise-perturbed counterparts, at a chosen epoch $\tau$.
To generate the noise-perturbed counterparts, we apply additive gaussian noise to the input.
Mathematically, we denote the capture of features from a desired layer $l$ as,
\begin{align}
\label{eq:features}
    &f(x_i) = \sigma(W x_i + b), \\ \notag 
    &f(x_i + \delta_i) = \sigma(W (x_i + \delta_i) + b).
\end{align}
Here, $\delta_i \sim \mathcal{N}(0,0.5)$ with dimensionality matching the input. 
Note: We drop the layer superscript to improve readability hereon. 
To avoid inconsistencies between the effects of applying $\delta_i$ independently at multiple layers, we apply $\delta_i$ to the image directly and observe its effects at downstream layers. 
Furthermore, to ensure that the noise is in the same feature space as the image, we apply the noise to the normalized image.

Once we obtain the features from each layer, we compute the distance $D(.)$  between corresponding pairs of features.
Here,
\begin{equation}
\label{eq:divergence}
\Delta f(i) = D(H(f(x_i)), H(f(x_i+\delta_i))) = \vert \vert H(f(x_i)) - H(f(x_i+\delta_i)) \vert \vert_2,
\end{equation}
where $H(.)$ is a projection function that maps the features into a lower dimensional space, and $\Delta f(i) \in \mathbb{R}^{1 \times O^{(l)}}$. 
The function $H: \mathbb{R}^{O^{(l)} \times h^{(l)} \times w^{(l)}} \rightarrow \mathbb{R}^{ O^{(l)} \times P}$, where $P << h^{(l)} \times w^{(l)}$ and $O^{(l)}$ denotes the filter counts from layer $l$.
While (\ref{eq:divergence}) depicts the $l_2$-norm version of the distance function, the formulation itself is not limited to it.
Beyond capturing the distance, we further normalize their values between samples to ensure that the distances remain comparable. 
We propose normalizing them on a channel-wise basis using the following equation,
\begin{equation}
    \label{eq:divergence_normalize}
    \Delta \hat{f}(i,m)= \frac{\Delta f(i,m)- \min\limits_{n \in 1,\ldots,N}\Delta f(n,1)}{\max\limits_{n \in 1,\ldots,N}\Delta f(n,1) - \min\limits_{n \in 1,\ldots,N}\Delta f(n,1)}.
\end{equation}
Here, $i \in \{1,2,\ldots,N\}$ and $m \in \{1,2,\ldots,O^{(l)}\}$.

\subsubsection{Sensitivity Constraint}
When collecting features across all the filters of a layer (\ref{eq:features}) we implicitly make the assumption of uniform importance across all filters.
However, from DNN pruning literature~\cite{DBLP:conf/iclr/0022KDSG17,yu2018nisp} we know that there are a number of filters which provide redundant information and reducing their contribution does not hurt the performance of DNNs.
Following this line of thought, we adopt the notion of sensitivity~\cite{ganesh2020slimming} to capture features from a subset of filters that provide important information.
While there are many different ways to combine sensitivity with the value of the features themselves, in this work we threshold the value of sensitivity to obtain a subset of the filters ($\tilde{O}^{(l)}$) from which we derive our features.
Doing so allows us to leverage the learned structure of the weight matrices in identifying sensitive filters while also reducing the overall memory consumed to store features.
The exact number of the subset of filters used for each DNN is provided in the supplementary materials.

\subsubsection{Computing the Binary Mask}
While $\Delta \hat{f}$ captures the distance between features from a specific layer, we expand the formulation of \alg{} to include the aggregation of distances across multiple layers of the DNN.
To do so, we include $\xi_i^{(l)}$, the instability of a sample measured as the average $\Delta \hat{f}$ across filters in a given layer. 
\begin{equation}
\label{eq:instability_layer}
\xi_i^{(l)} = \frac{\sum_{m=1}^{\tilde{O}^{(l)}} \Delta \hat{f}(i,m)}{\tilde{O}^{(l)}}, \;\; i \in \{1,\ldots,N\}.
\end{equation}
By combining the contributions of $\xi_i^{(1)}, \xi_i^{(2)}, \ldots, \xi_i^{(L)}$ across multiple layers we obtain the overall instability of a sample, $\xi_i$, given as,
\begin{equation}
\label{eq:instability_sample}
\xi_i = \xi^{(1)}_i \alpha^{(1)} + \ldots + \xi^{(L)}_i \alpha^{(L)},
\end{equation}
where $\alpha()$ denotes a window function that provides scalar multipliers used to combined the instability values obtained from different layers.

To identify the optimal values of $\alpha$ would require solving the system of equations shown below,
\begin{equation}
\begin{bmatrix}
\xi_1^{(1)} & \xi_1^{(2)} & \ldots & \xi_1^{(L)} \\
\vdots & \vdots & \ldots & \vdots \\
\xi_N^{(1)} & \xi_N^{(2)} & \ldots & \xi_N^{(L)} 
\end{bmatrix} \in \mathbb{R}^{(N \times L)}
\begin{bmatrix}
\alpha(1)\\
\vdots \\
\alpha(L) 
\end{bmatrix} \in \mathbb{R}_{\geq 0}^{(L)},
\end{equation}
where the final accuracy is the metric over which we need to optimize.
Given the practical constraints in solving this system of equations, where the LHS is ill-defined and the size of the system matrix forces any operation on it to be expensive, we explore a restricted set of functions, including an $\mathbbm{1}_{1:\frac{L}{2}}$, $\mathbbm{1}_{\frac{L}{2}:L}$, a gaussian distribution and finally $\mathbbm{1}_L$, to find the best performing $\alpha$.  
Once we set $\alpha$, we can evaluate $\xi_i$. 
Using these values, $m$ can be computed as:
\begin{equation} 
    m_i = \begin{cases} 
      0 & \text{if}~ \xi_i ~\text{is in the top}~ \gamma ~\text{values of}~ \xi \\
      1 & \text{o.w} \enspace . 
   \end{cases}
\end{equation}
\noindent By controlling $\gamma$, we use $m_i$ to reduce the amount of the training data held in memory as well as the overall training time required.
Once $m$ is applied, the DNN is then trained with the remaining subset of data from epochs $\tau$ to $E$.
\section{Experimental Results}
\label{sec:experimental_results}

The experimental results section is divided into three main parts with each aligning with one PER goal. The first discusses the performance of \alg{} in the context of the state-of-the-art curriculum learning algorithm~\cite{zhou2020curriculum}. 
The second part emphasizes the adversarial robustness of \alg{}, in the context of normal as well as adversarial training, under a variety of adversarial attacks.  The third part demonstrates the improvement in efficiency.

\subsection{Setup}
\label{subsec:setup}
We briefly outline the datasets, DNNs, types of adversarial attacks and metrics used across our experiments. 
We provide details of the hyper-parameters and experimental setups in the supplementary materials.

\paragraph{Datasets} 
We use five primary datasets to evaluate our proposed method, CIFAR-10, CIFAR-100~\cite{krizhevsky2009learning}, STL-10~\cite{coates2011analysis},  miniImagenet~\cite{vinyals2016matching} and ILSVRC2012~\cite{ILSVRC15}. 
Among these datasets, we restrict our adversarial robustness comparisons to CIFAR-10 to ensure that we adhere to the page limit. 
For miniImagenet, we use a custom-generated and balanced training-and-testing split that we will make available alongside our code.

\paragraph{DNN architectures} 
We use four DNN architectures to evaluate \alg{}, VGG16 \cite{DBLP:journals/corr/SimonyanZ14a}, MobileNet~\cite{DBLP:conf/cvpr/SandlerHZZC18}, DenseNet~\cite{huang2019convolutional,huang2017densely} and ResNet50~\cite{DBLP:conf/cvpr/HeZRS16}. 
These networks were chosen with a view to represent a wide variety of architectures.
Each DNN has two distinct versions, one suitable for the CIFAR datasets and another for the remaining datasets.\footnote{Detailed descriptions of these model variants are provided in our code base.  The link for the repository will be here upon camera ready publication.}

\paragraph{Adversarial Attacks And Metrics}
We explore the effect of a variety of adversarial attacks like MIFGSM \cite{DBLP:conf/cvpr/DongLPS0HL18}, FFGSM~\cite{DBLP:conf/iclr/WongRK20}, DI2FGSM~\cite{DBLP:conf/cvpr/XieZZBWRY19}, APGDDLR~\cite{DBLP:conf/icml/Croce020a}, APGDCE, PGD~\cite{madry2018towards} and CW~\cite{DBLP:conf/sp/Carlini017} using the code from ~\cite{kim2020torchattacks,zhang2020attacks}. 
To measure the performance of various algorithms, we use standard \textbf{Accuracy~($\%$)} over the testing set. For adversarial robustness we measure \textbf{Accuracy~($\%$)} over the perturbed testing set, illustrated by the radius of the polar plots. Finally, we use total \textbf{Training Time (minutes)} to compare the improvement in efficiency across different training methods.
Across all experiments, we provide average statistics over 5 trials unless stated otherwise.

\subsection{Curriculum Comparison}
\label{subsec:curriculum_comparison}

\begin{table}[t!]
\centering
\caption{Comparison of \alg{} ($\tau = 50$, $\alpha=\mathbbm{1}_L$) with mini-batch SGD, DIHCL and Random baselines. Across most datasets \alg{} achieves the best performance. Here, \textbf{bold} refer to the best performance while \underline{underline} refer to the second best method.* indicates numbers cited by authors. -- indicate combinations  we avoided due to limited run-time. ILSVRC2012 results are across 1 trial. Number of samples removed from the training set ($\gamma$) for each case is provided in the supplementary materials}
\label{tab:curriculum_comparison}
\begin{tabular}{@{}ccccccc@{}}
\toprule
DNN                    & Algorithm     & CIFAR-10 & CIFAR-100 & STL-10 & miniImagenet & ILSVRC2012\\ \midrule
\multirow{4}{*}{VGG16} & Baseline      & \underline{94.04}    & \underline{74.23}     & \underline{82.75}  & \underline{70.95} & --       \\
                       & Random        & 93.19    & 71.63     & 80.38  & 67.57 & --       \\
                       & DIHCL         & 94.03    & 72.89     & 79.71  & 66.07 & --       \\
                       & Q-TART (Ours) & \textbf{94.47}     & \textbf{75.06} & \textbf{83.01}  & \textbf{71.61}  & --      \\ \midrule 
\multirow{4}{*}{MobileNet} & Baseline      & \underline{93.50}    & \underline{72.75}     & \underline{77.95}  & \underline{64.62}      & --  \\
                           & Random        & 92.31    & 71.15     & 73.86  & 62.11 & --       \\
                           & DIHCL         & 88.97    & 61.58     & 75.40  & 49.37 & --       \\
                           & Q-TART (Ours) & \textbf{93.62} & \textbf{74.97} & \textbf{80.04} & \textbf{66.92}  & --\\ \midrule 
\multirow{4}{*}{DenseNet} & Baseline       & \underline{95.13}    & \underline{76.95}     & 85.55  & \underline{73.78}        & --\\
                           & Random        & 93.88    & 74.18     & 82.39  & 71.23       & -- \\
                           & DIHCL         & 94.72    & 76.03     & \underline{85.82}  & 64.34 & --        \\
                           & Q-TART (Ours) & \textbf{95.19} & \textbf{77.74} & \textbf{85.83} & \textbf{75.97} & --\\ \midrule 
\multirow{4}{*}{ResNet50}  & Baseline      & 95.63    & 79.27     & 72.77  & \underline{68.76}  & 76.32      \\
                           & Random        & 95.27    & 76.71     & 69.29  & 64.69  & --      \\
                           & DIHCL         & \textbf{95.83}    & \underline{79.71}     & \textbf{73.58}  & 66.86        & \underline{76.33}*\\
                           & Q-TART (Ours) & \underline{95.75} & \textbf{79.78} & \underline{73.40} & \textbf{69.77} & \textbf{77.04}\\ \bottomrule 
\end{tabular}
\end{table}

In this experiment, our main goal is to compare the performance of \alg{} against mini-batch SGD training and highlight how we can improve performance while only retaining a subset of our training data.
Additionally, we compare against the state-of-the-art curriculum learning method DIHCL~\cite{zhou2020curriculum} that prioritizes the removal of samples throughout the training process.
We extend their code to accommodate our datasets and DNN architectures while maintaining their training protocols. 

From Table.~\ref{tab:curriculum_comparison}, across all combinations of datasets and DNN architectures, we observe that our algorithm easily outperforms the baseline mini-batch SGD setup, even with the removal of a subset of the training data.
To ensure fair comparison, we used the exact same hyper-parameter setups across both methods.
More interestingly, when we observe the performance of DIHCL adapted to our selection of dataset-DNN pairs we see that it consistently exhibits strong performances on the ResNet architectures.
This, in conjunction with DIHCL's propensity to perform significantly worse than randomly removing the same number of samples as in \alg{} (marked in Table as Random) across the other tested architectures may point towards the strong affinity of the training setup used in DIHCL to residual architectures.
However, even with the starkly different training setup used in DIHCL (which includes cyclic learning rate schedules, a teacher-like copy of the DNN, etc.), \alg{} outperforms it in most cases.

\subsection{Ablation: Window Functions}
\label{subsec:ablation_window_functions}

\begin{table}[t!]
\centering
\caption{Assessing the susceptibility of samples to noise using multiple layers boosts $\gamma$ as well as Accuracy~($\%$). $\alpha = \mathbbm{1}_{1:\frac{L}{2}}$ provides the best performance. Optimal result for each CIFAR10-DNN-$\alpha$ combination is provided in the table}
\label{tab:window}
\begin{tabular}{@{}cllll@{}}
\toprule
Algorithm     & VGG16 & MobileNet & DenseNet & ResNet50 \\ \midrule
Baseline   & 94.04    & 93.50     & 95.13  & 95.63        \\
$\alpha=\mathbbm{1}_L$         & 94.47 ($\gamma=125$)& 93.62 ($\gamma=125$) & 95.19 ($\gamma=50$)  & 95.75 ($\gamma=25$)\\
$\alpha=\mathbbm{1}_{1:\frac{L}{2}}$     & \textbf{94.49} ($\gamma=300$) & \textbf{93.66} ($\gamma=150$) & \textbf{95.28} ($\gamma=50$)  & \textbf{95.78} ($\gamma=50$)\\
$\alpha=\mathbbm{1}_{\frac{L}{2}:L}$     & 94.41 ($\gamma=300$) & 93.59 ($\gamma=150$) & 95.22 ($\gamma=50$)  & 95.72 ($\gamma=50$)\\
$\alpha=\mathcal{N}(0,1)$      & 94.43 ($\gamma=250$) & 93.61 ($\gamma=125$) & 95.16 ($\gamma=300$)  & 95.74 ($\gamma=100$)\\
\bottomrule
\end{tabular}
\end{table}

Across all the results presented in Table~\ref{tab:curriculum_comparison} we assume the use of $\alpha = \mathbbm{1}_L$, which results in the collection of features from the last convolutional layer.
In this section, we compare and contrast four different window functions to identify the best performing function.
Based on Table~\ref{tab:window}, there are two main observations.
First, the use of additional layers in assessing the susceptibility of samples to noise often allows for an increase in $\gamma$ when compared to the case of $\mathbbm{1}_L$, with a minor trade-offs in performance. 
Second, in conjunction with the first observation, $\alpha = \mathbbm{1}_{1:\frac{L}{2}}$ shows the best Accuracy~($\%$) across our restricted set of window functions. 
These results highlight the regularization effect our method imposes on the DNN, regardless of the location at which we ascertain the distance between features.
Further, by assessing distances across layers other than the final one in the DNN, we reduce the relationship between specific task-oriented information and how we assess noisy samples, allowing \alg{} to be more extensible to alternative tasks. 

\subsection{Adversarial Robustness}
\label{subsec:adversarial_robustness}

We discuss the robustness of \alg{} to a variety of adversarial attacks on CIFAR-10.
We measure adversarial robustness to attacks where the source and target models are the same as well as the case when multiple source models are used to generate the attacks for a single target.
To ensure parity, we avoid comparing results across methods with and without adversarial training.

\subsubsection{Same Adversarial Source and Target}
\label{subsubsec:source_equal_target}
\begin{figure}[t!]
\centering
\subfloat{\includegraphics[width=0.3\columnwidth]{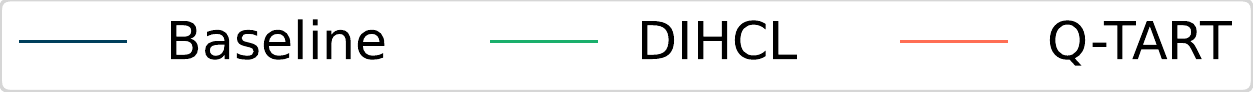}}

\subfloat[][VGG16]{\includegraphics[width=0.24\columnwidth]{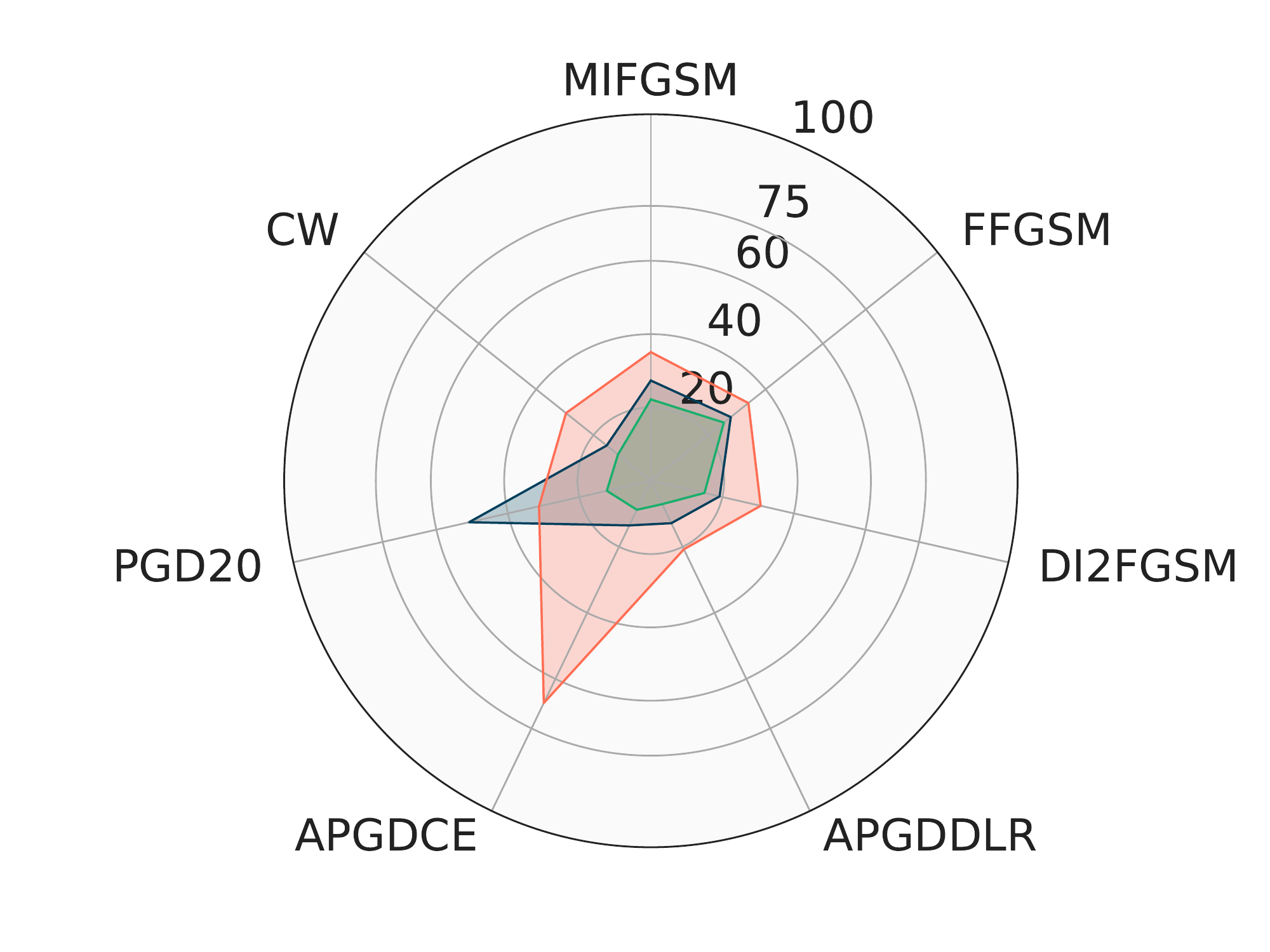} \label{fig:noat_vgg16}}
\subfloat[][MobileNet]{\includegraphics[width=0.24\columnwidth]{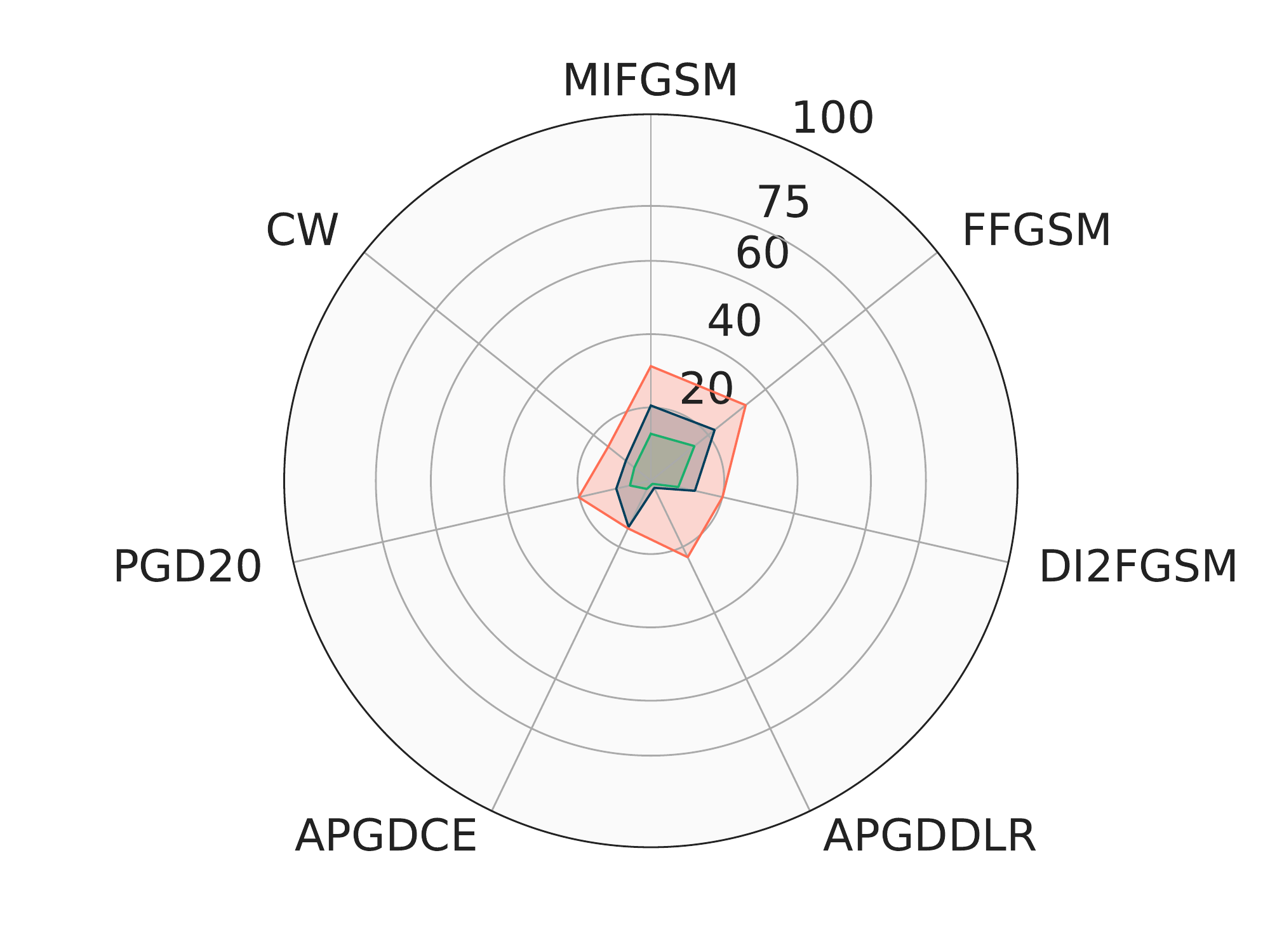} \label{fig:noat_mobilenet}}
\subfloat[][DenseNet]{\includegraphics[width=0.24\columnwidth]{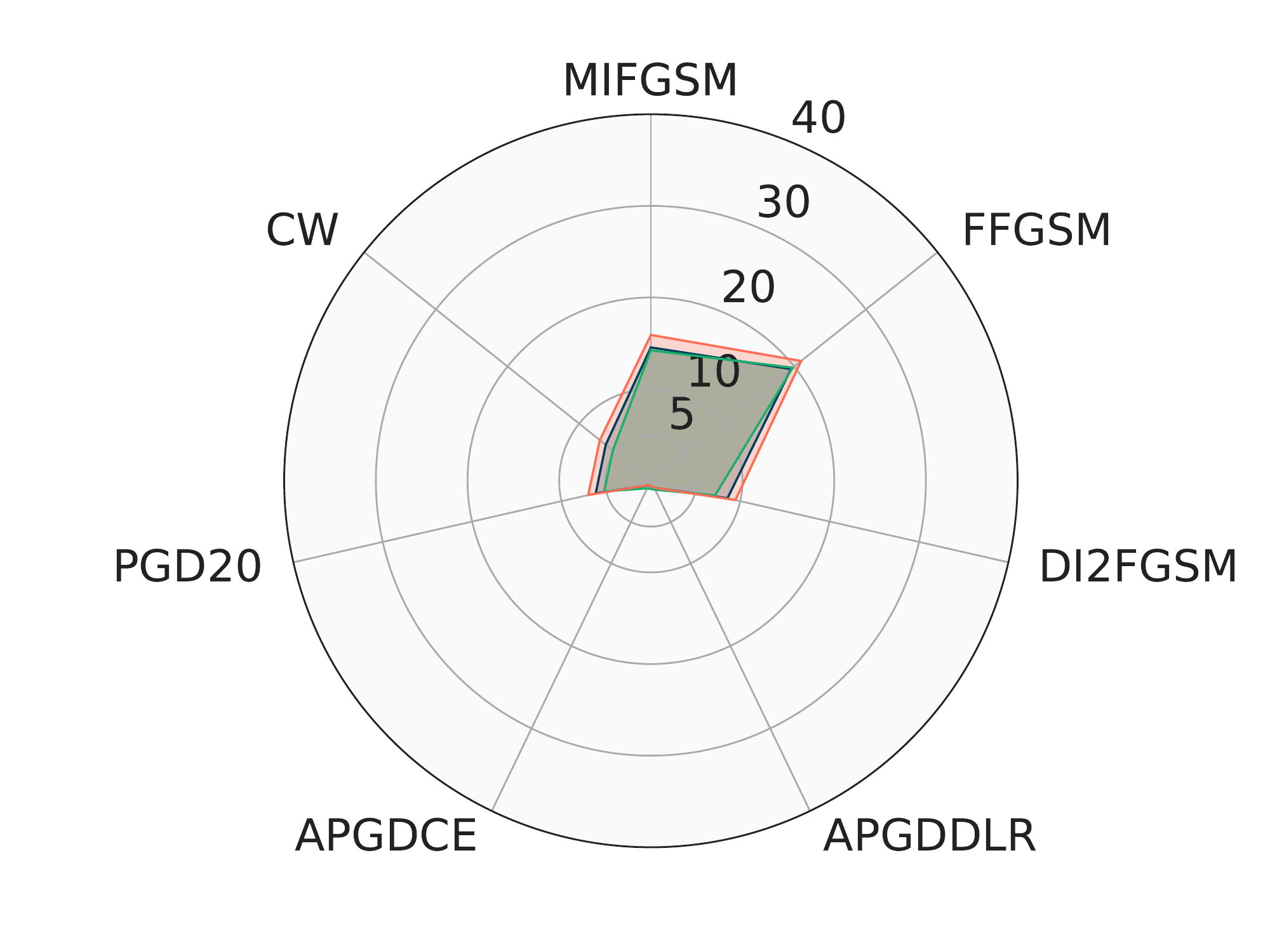} \label{fig:noat_densenet}}
\subfloat[][ResNet50]{\includegraphics[width=0.24\columnwidth]{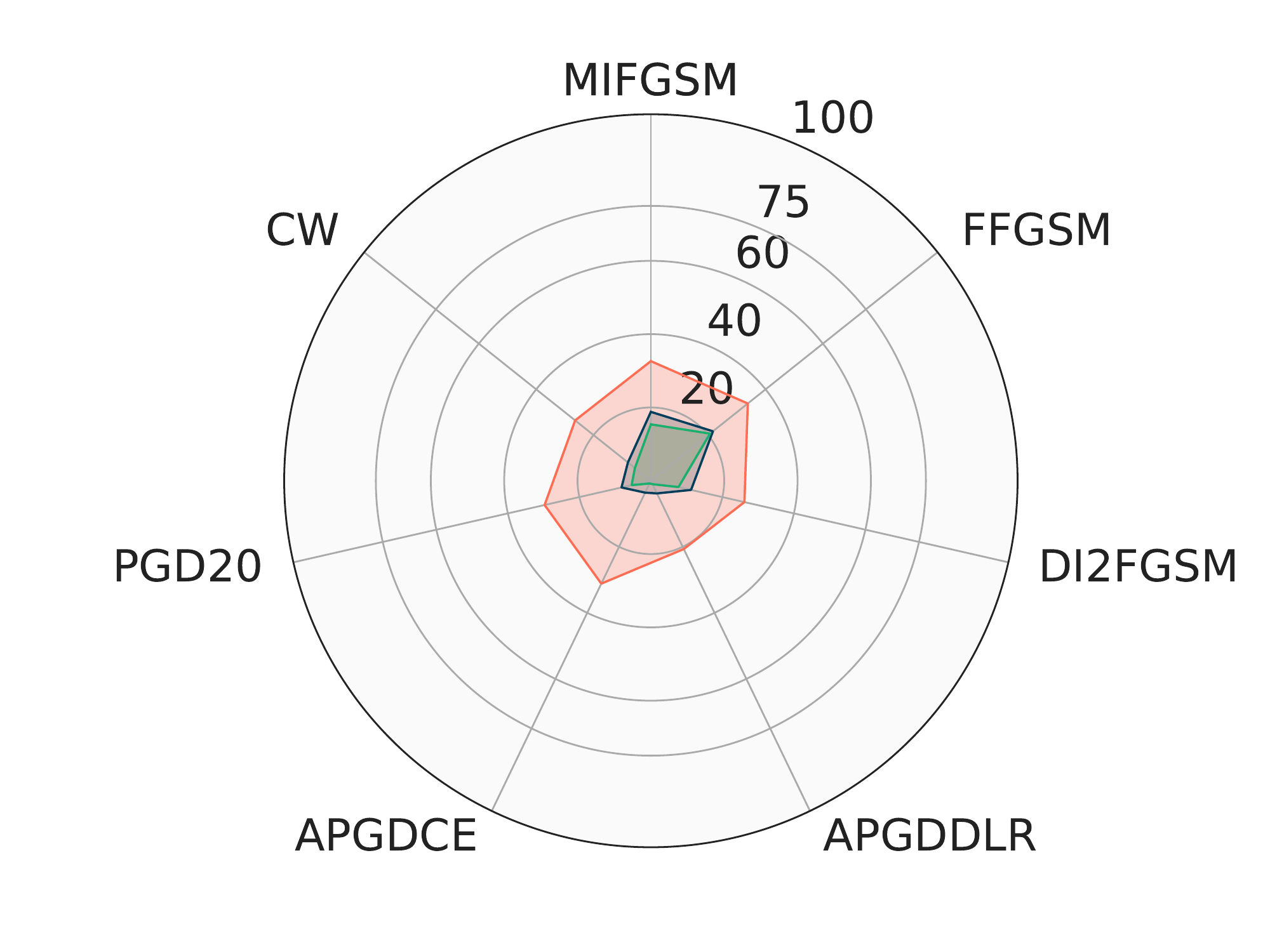} \label{fig:noat_resnet}}
\caption{Across all DNN architectures, \alg{} matches and often significantly improves upon the adversarial robustness of mini-batch SGD training and DIHCL. Methods with the largest area of plot are preferred. $\gamma$ values are 125, 125,50 and 25 for VGG16, MobileNet, DenseNet and ResNet50 respectively}
\label{fig:no_at}
\end{figure}

Using Fig.~\ref{fig:no_at} we establish two main observations, 1) in multiple instances DIHCL reduces the robustness of DNNs when compared to mini-batch SGD training, and more importantly 2) \alg{} significantly improves the robustness of DNNs to multiple adversarial attacks, with DenseNet showing the lowest improvement overall.
It is important to highlight the fact that we recover and remove noisy samples permanently in \alg{} without making an explicit assumption on the adversarial nature of the noise being used including, fine-tuning the perturbations added to the image or a change in predictions from the DNN.

\begin{figure}[ht!]
\centering
\subfloat{\includegraphics[width=0.6\columnwidth]{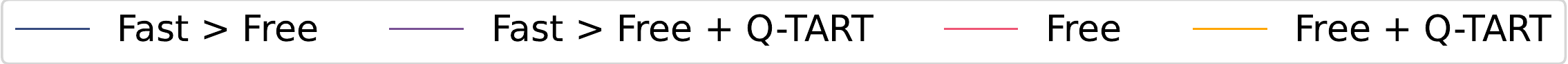}}

\subfloat[][VGG16]{\includegraphics[width=0.24\columnwidth]{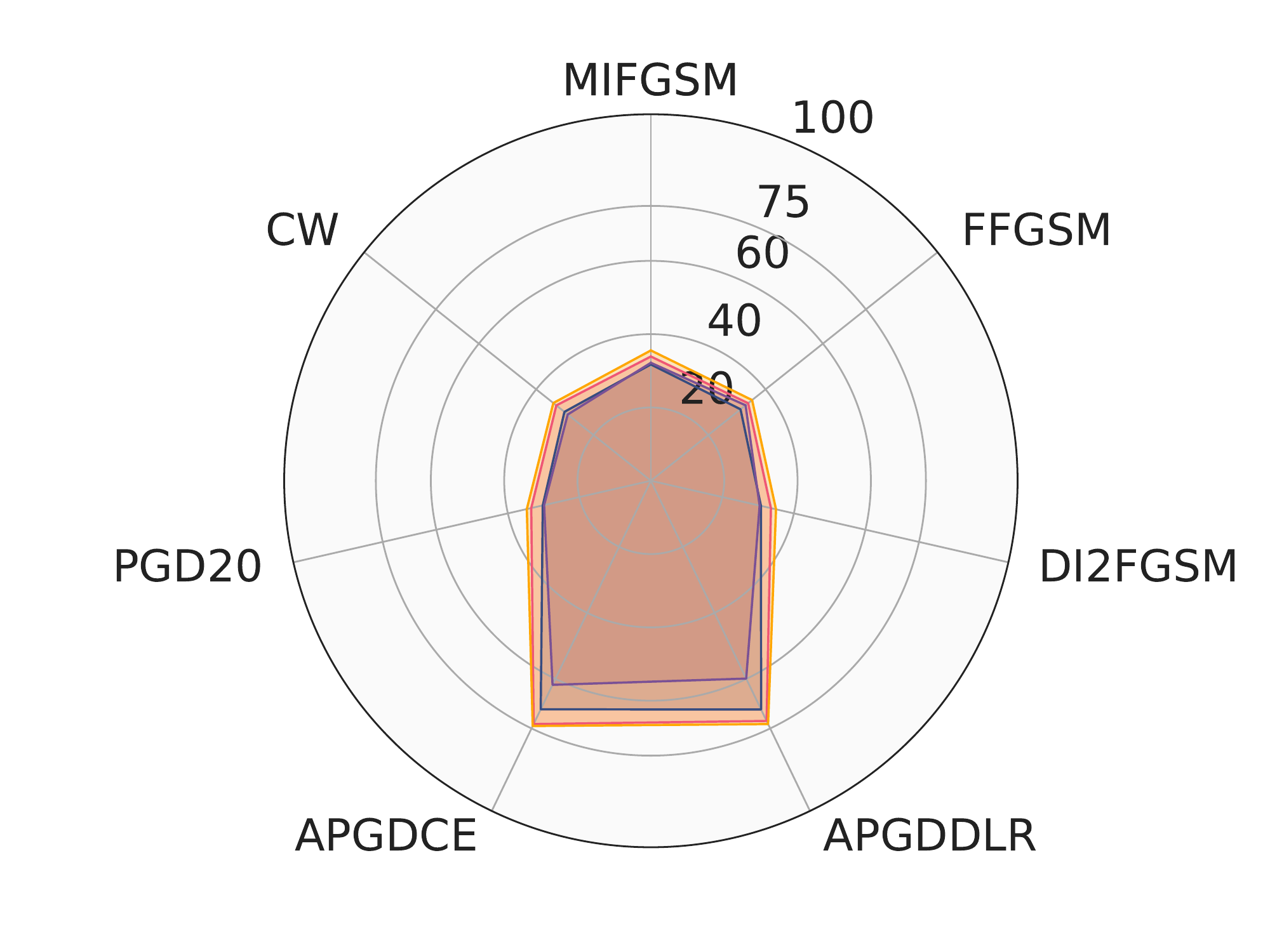} \label{fig:at_vgg16}}
\subfloat[][MobileNet]{\includegraphics[width=0.24\columnwidth]{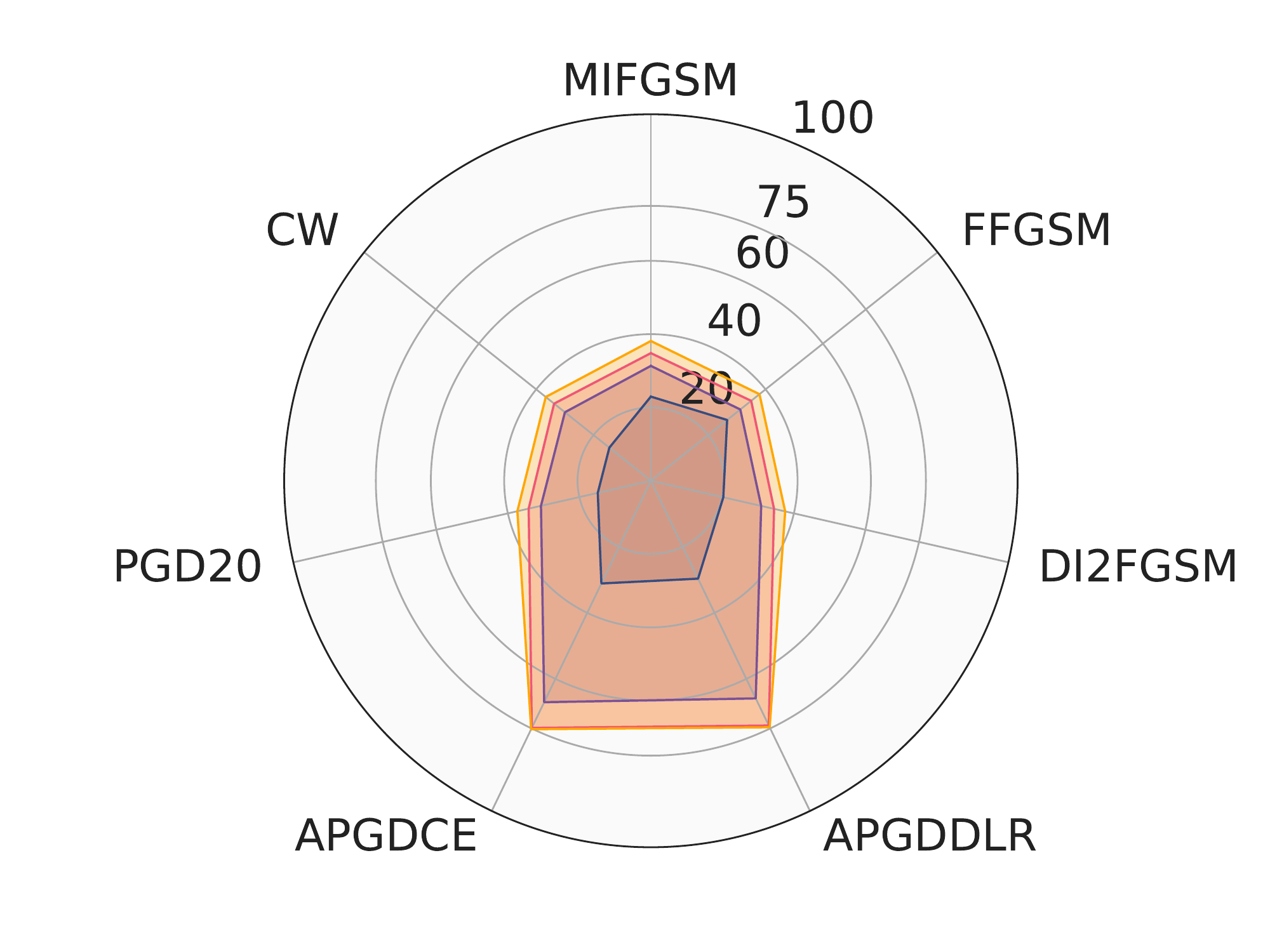} \label{fig:at_mobilenet}}
\subfloat[][DenseNet]{\includegraphics[width=0.24\columnwidth]{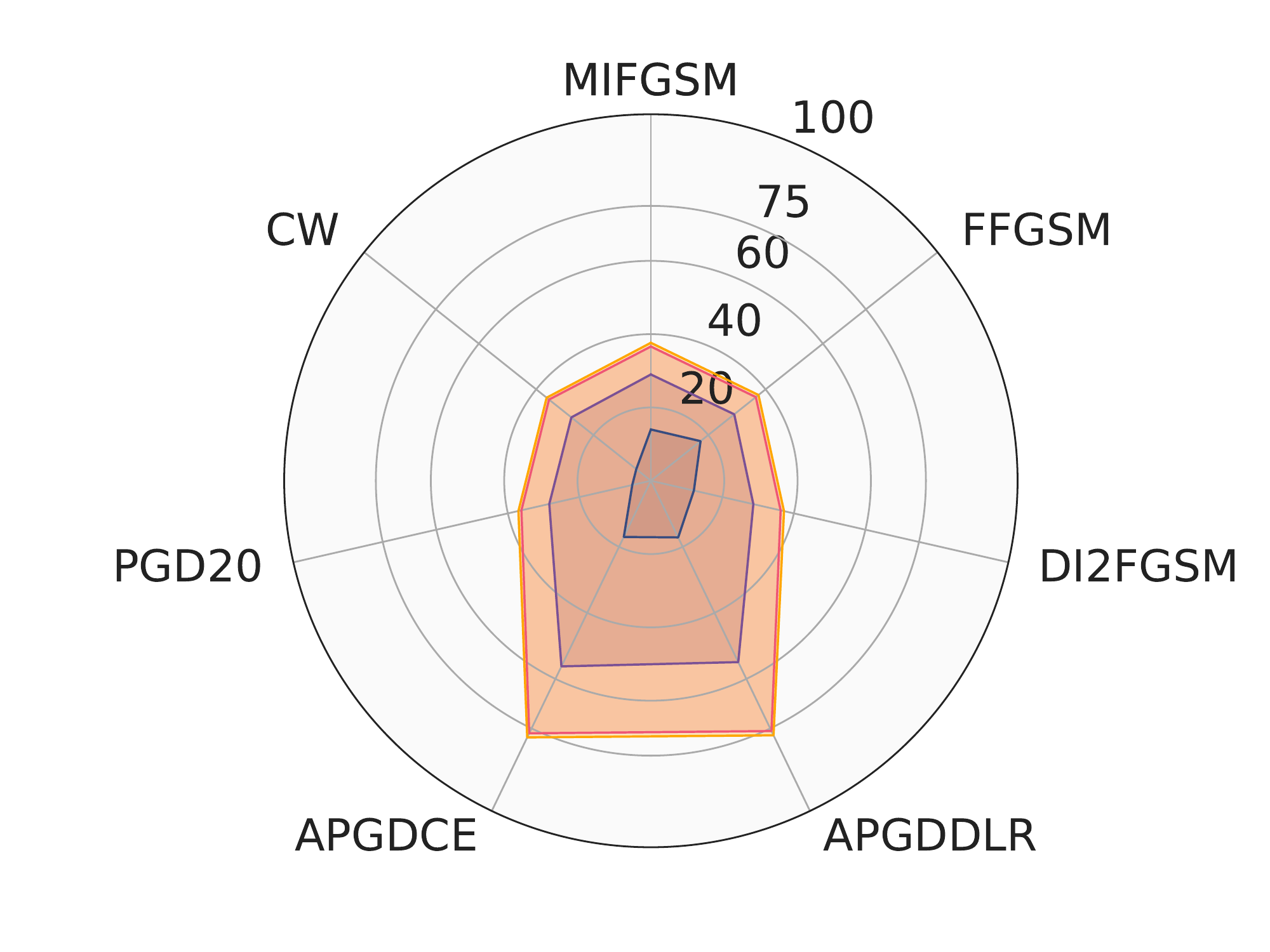} \label{fig:at_densenet}}
\subfloat[][ResNet50]{\includegraphics[width=0.24\columnwidth]{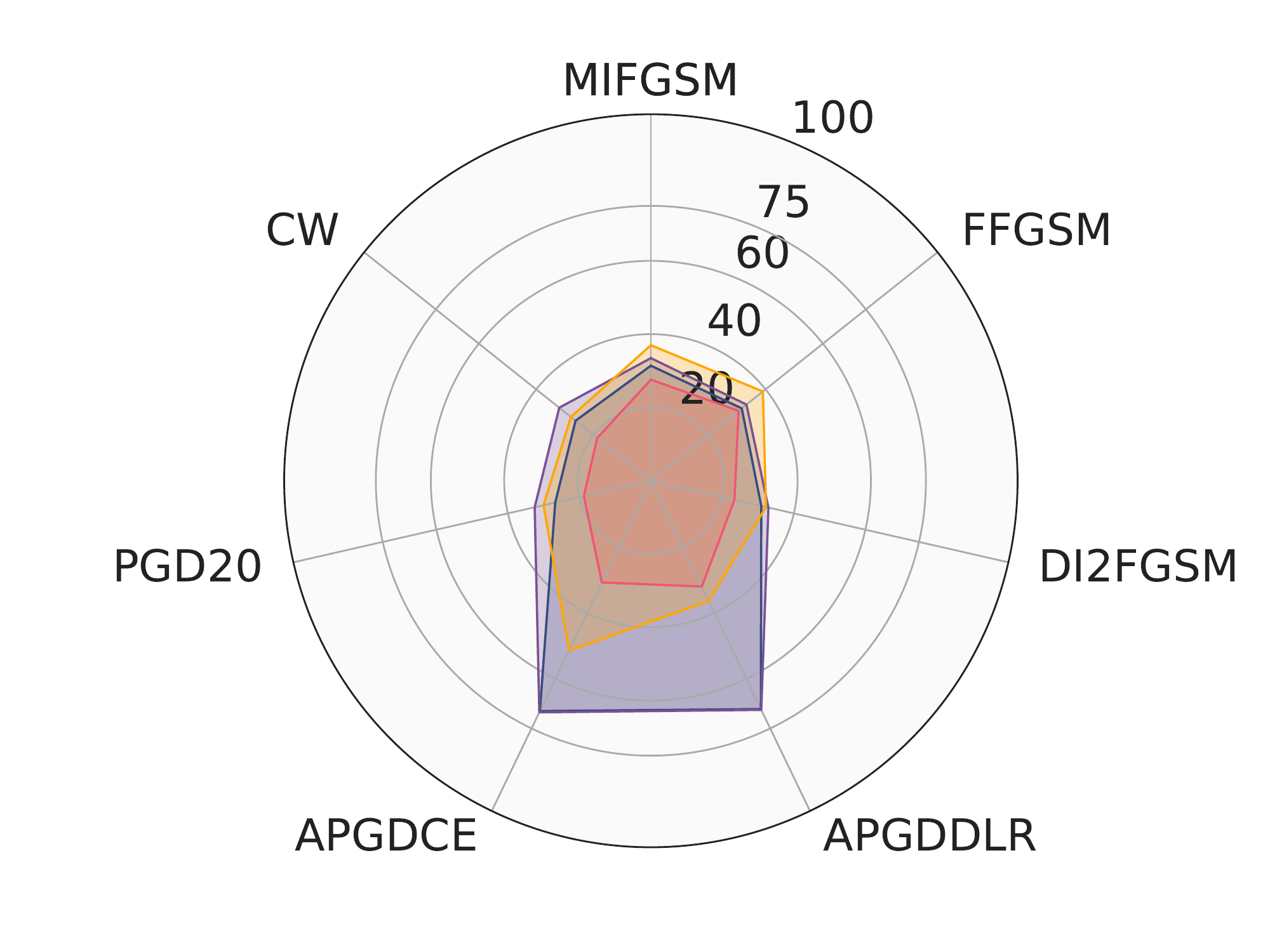} \label{fig:at_resnet}}
\caption{Largest improvements from adversarial training are observed for PGD, CW, APGDDLR and APGDCE attacks. More generally, the addition of \alg{} atop adversarial training methods improves the robustness of the DNN. Results provided for methods using \cite{shafahi2019adversarial} are across 1 trial. $\gamma$ values are 125/12, 125/125, 50/5 and 12/25 for \alg{} + \cite{DBLP:conf/iclr/WongRK20} and \alg{} + \cite{shafahi2019adversarial} respectively across VGG16, MobileNet, DenseNet and ResNet50}
\label{fig:at}
\end{figure}

Adversarial training approaches impart robustness to DNN by exposing them to multiple examples of adversarial input during the training phase.
In general, adversarial training seems to improve performance on PGD20, CW, APGDDLR and APGDCE attacks when compared to the most robust performance offered by \alg{}.
However, adding \alg{} atop common adversarial training approaches further boosts their performance against adversarial attacks, as shown in Fig.~\ref{fig:at}.
This increase in performance supports our hypothesis that \alg{} is complementary to adversarial training.
While \alg{} removes noisy samples from the training set itself, to build a more cohesive training set, it does not harm the robustness offered by exposing DNNs to various examples of adversarial input.
Interestingly, this supports the notion of an optimal training subset~\cite{DBLP:journals/corr/LapedrizaPBT13} during adversarial as well as normal mini-batch SGD training.

\subsubsection{Multiple Adversarial Sources Same Target}
\label{subsubsec:adversarial_source_notequal_target}

To measure in-Transferability, we use the mean and standard deviation of Accuracy~($\%$) when a selected model is attacked using adversaries generated from all four of the DNN architectures used in our experiments.
We specifically demand that standard deviation in performance is minimized, in addition to high average performance, since a high deviation is indicative of robustness being dependent on the type of DNN backbone used to generate adversaries. 
However, since our main assumption is that the adversaries can be generated from any possible source a lower standard deviation is preferred.

\begin{figure}[t!]
\centering
\subfloat{\includegraphics[width=0.6\columnwidth]{images/TRIAL_AT.pdf}}

\subfloat[][VGG16]{\includegraphics[width=0.24\columnwidth]{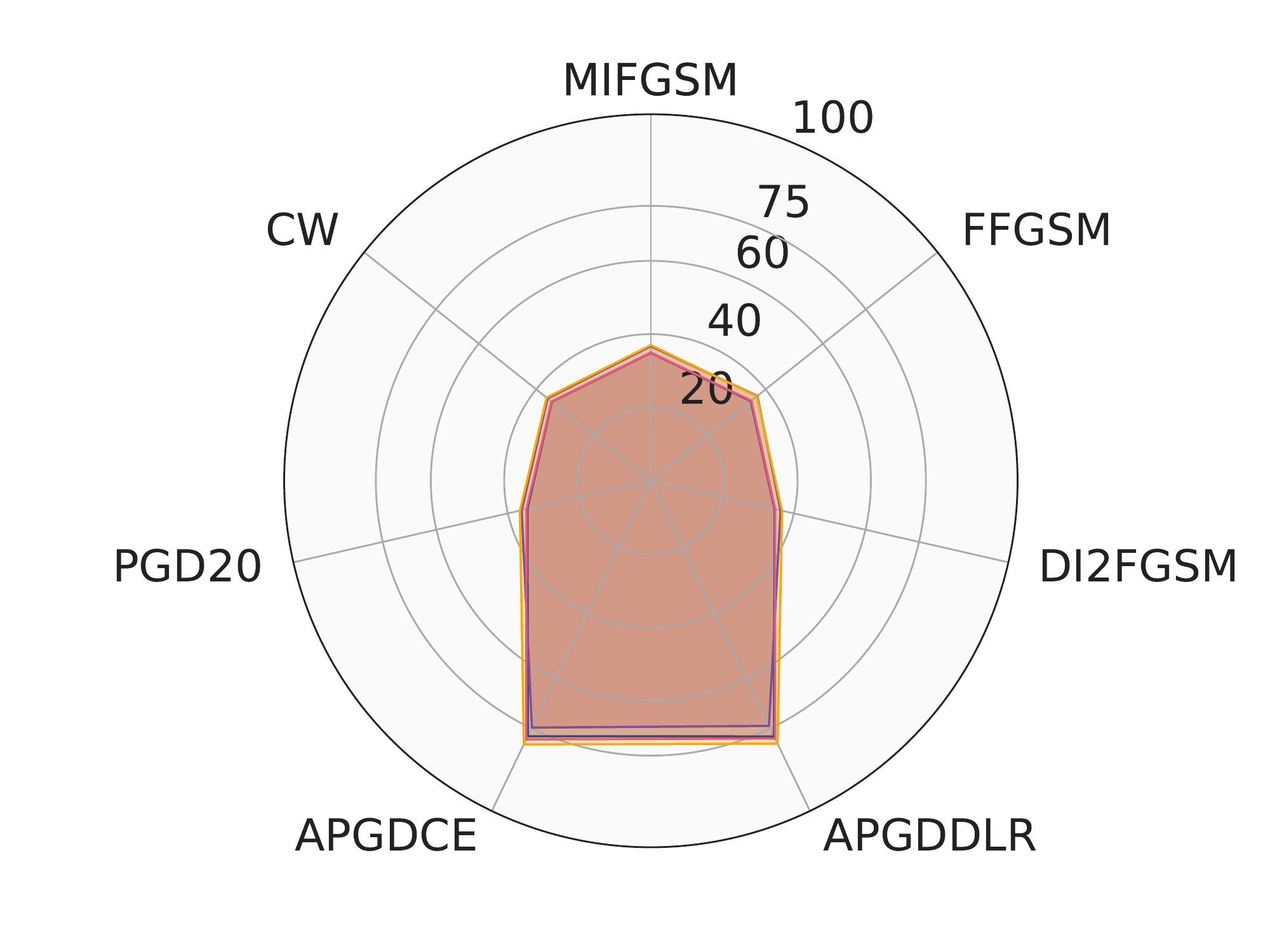} \label{fig:transfer_avg_vgg16}}
\subfloat[][MobileNet]{\includegraphics[width=0.24\columnwidth]{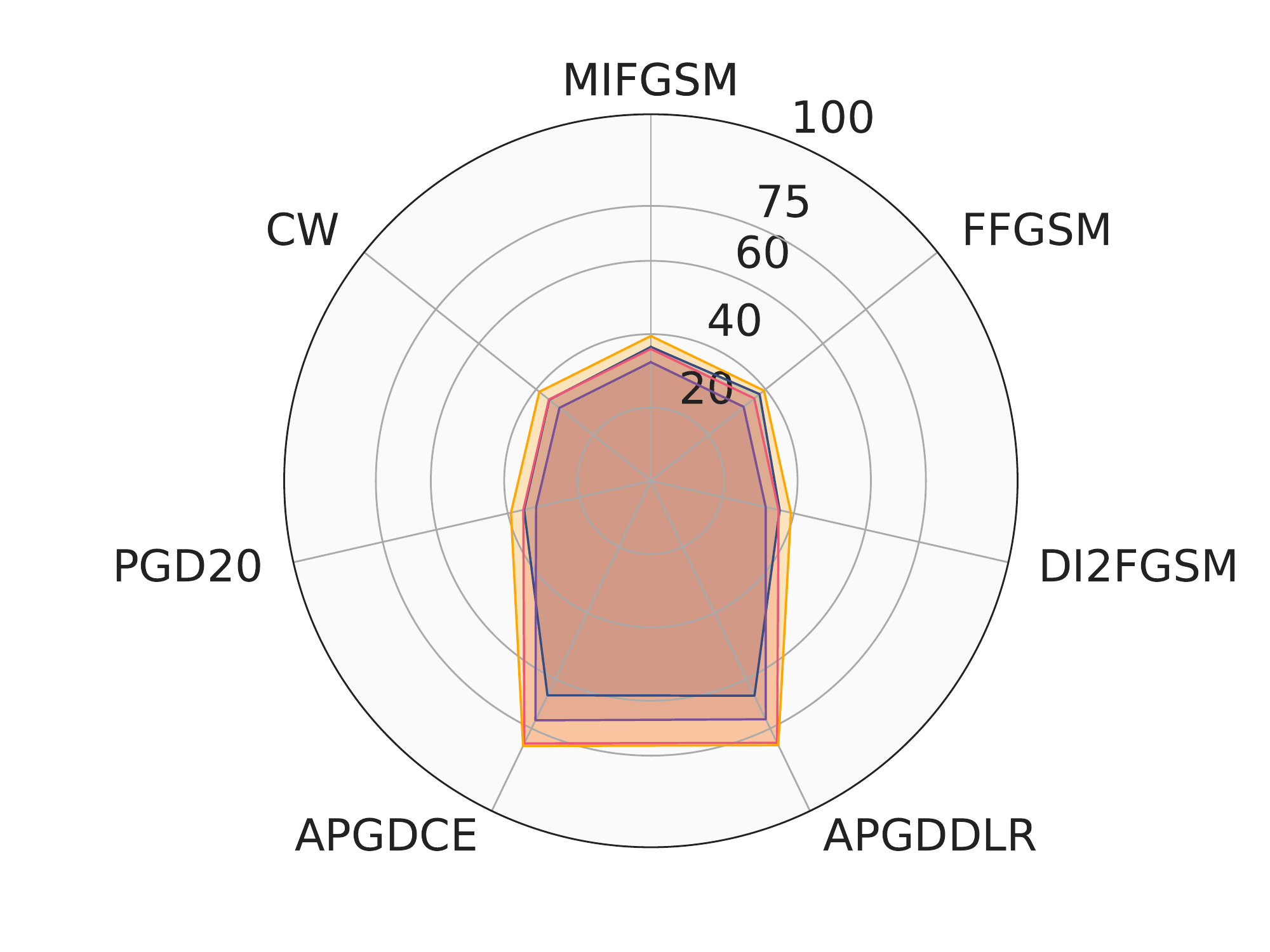} \label{fig:transfer_avg_mobilenet}}
\subfloat[][DenseNet]{\includegraphics[width=0.24\columnwidth]{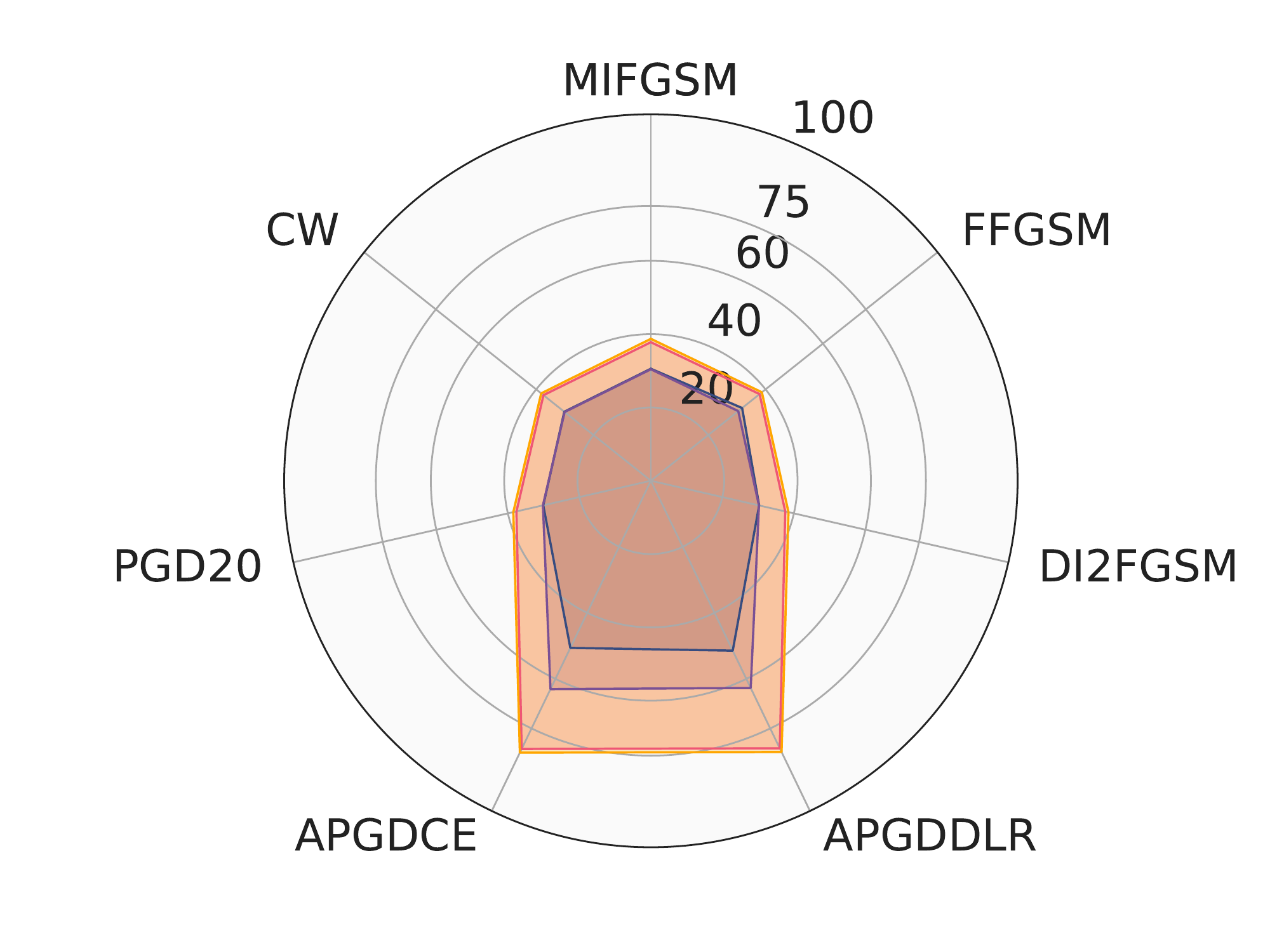} \label{fig:transfer_avg_densenet}}
\subfloat[][ResNet50]{\includegraphics[width=0.24\columnwidth]{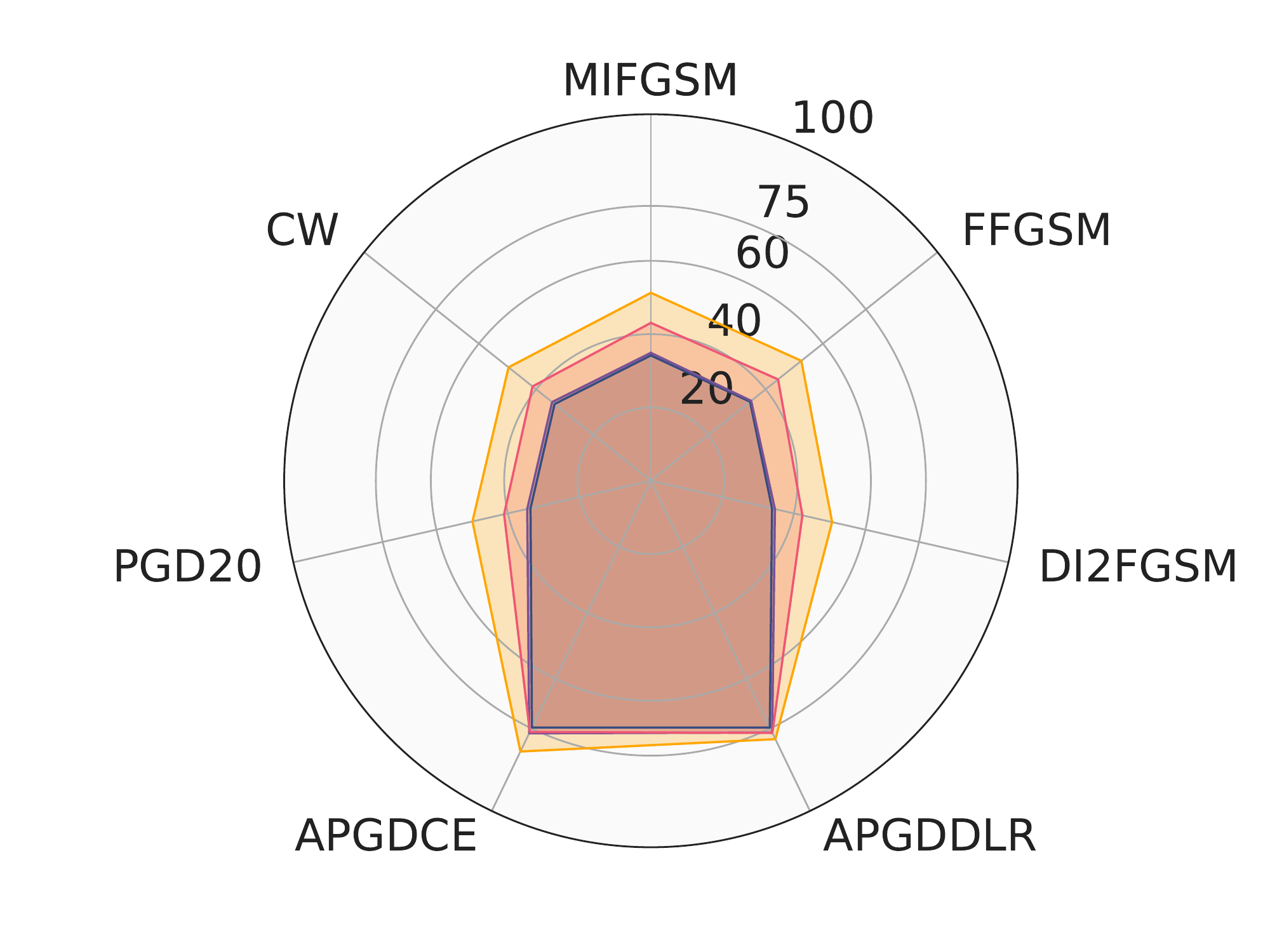} \label{fig:transfer_avg_resnet}}
\caption{Illustration of \alg{}-based training boosting the average robustness accuracy when tested across multiple sources of adversaries. In our experiments, we use all four possible DNN architectures to generate attacks. $\gamma$ values are 125/12, 125/125, 50/5 and 12/25 for \alg{} + \cite{DBLP:conf/iclr/WongRK20} and \alg{} + \cite{shafahi2019adversarial} respectively across VGG16, MobileNet, DenseNet and ResNet50}
\label{fig:avg_transfer}
\end{figure}

\begin{figure}[t!]
\centering
\subfloat{\includegraphics[width=0.6\columnwidth]{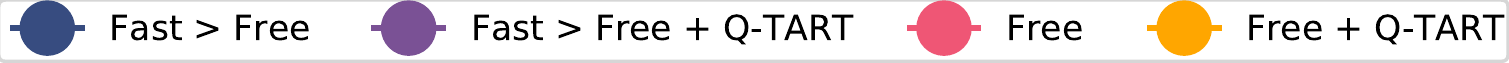}}

\subfloat[][VGG16]{\includegraphics[width=0.24\columnwidth]{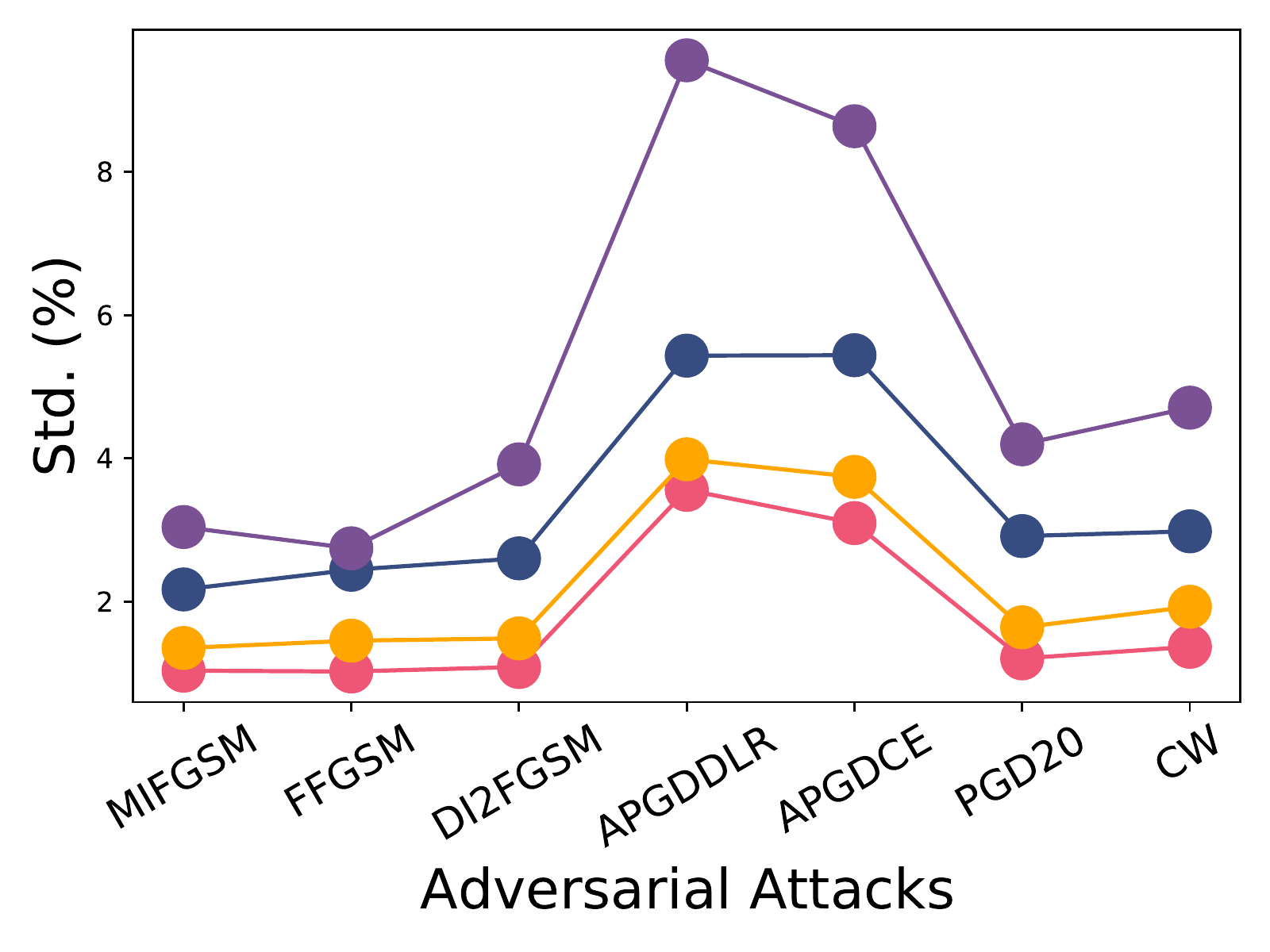} \label{fig:transfer_dev_vgg16}}
\subfloat[][MobileNet]{\includegraphics[width=0.24\columnwidth]{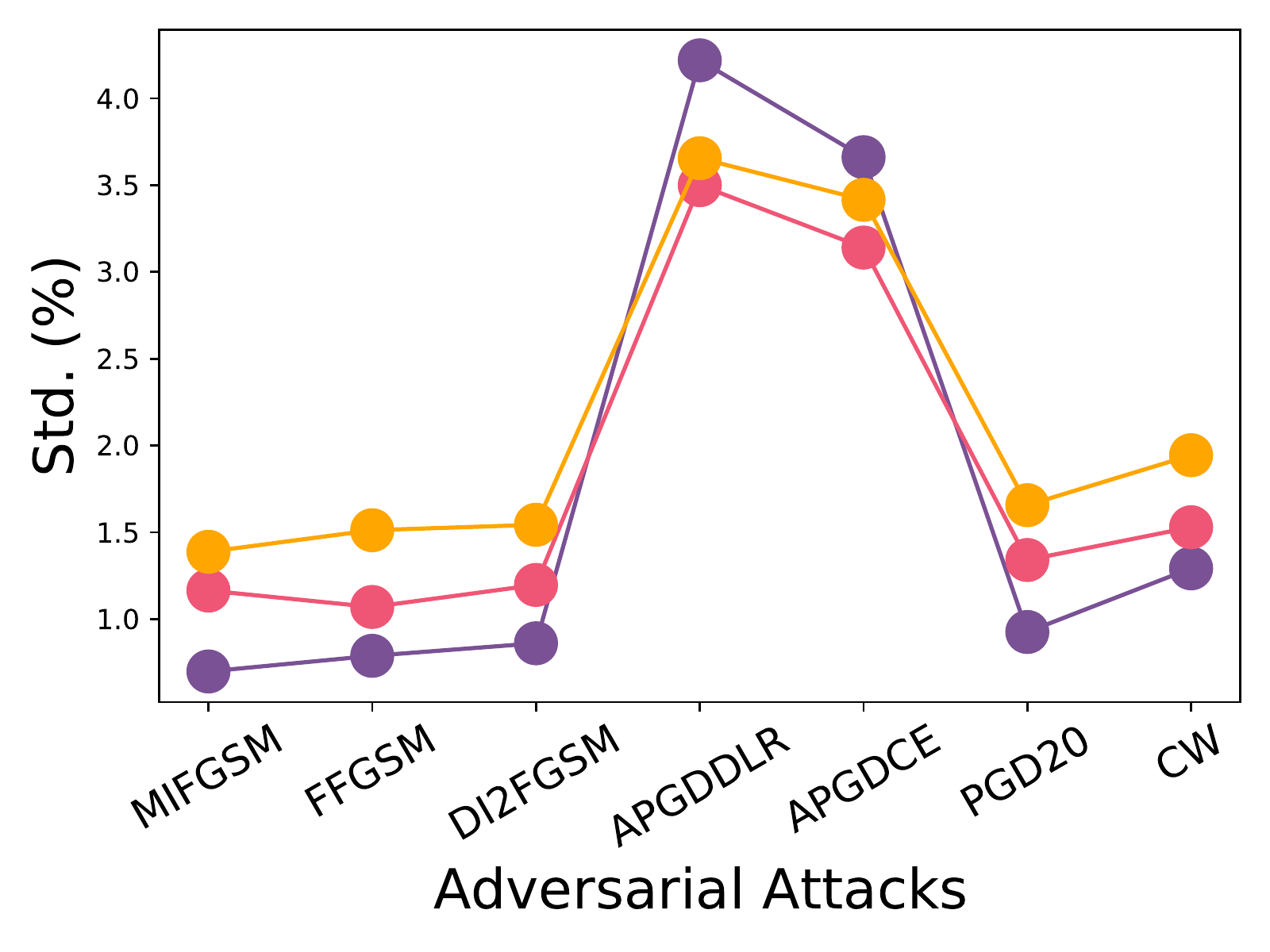} \label{fig:transfer_dev_mobilenet}}
\subfloat[][DenseNet]{\includegraphics[width=0.24\columnwidth]{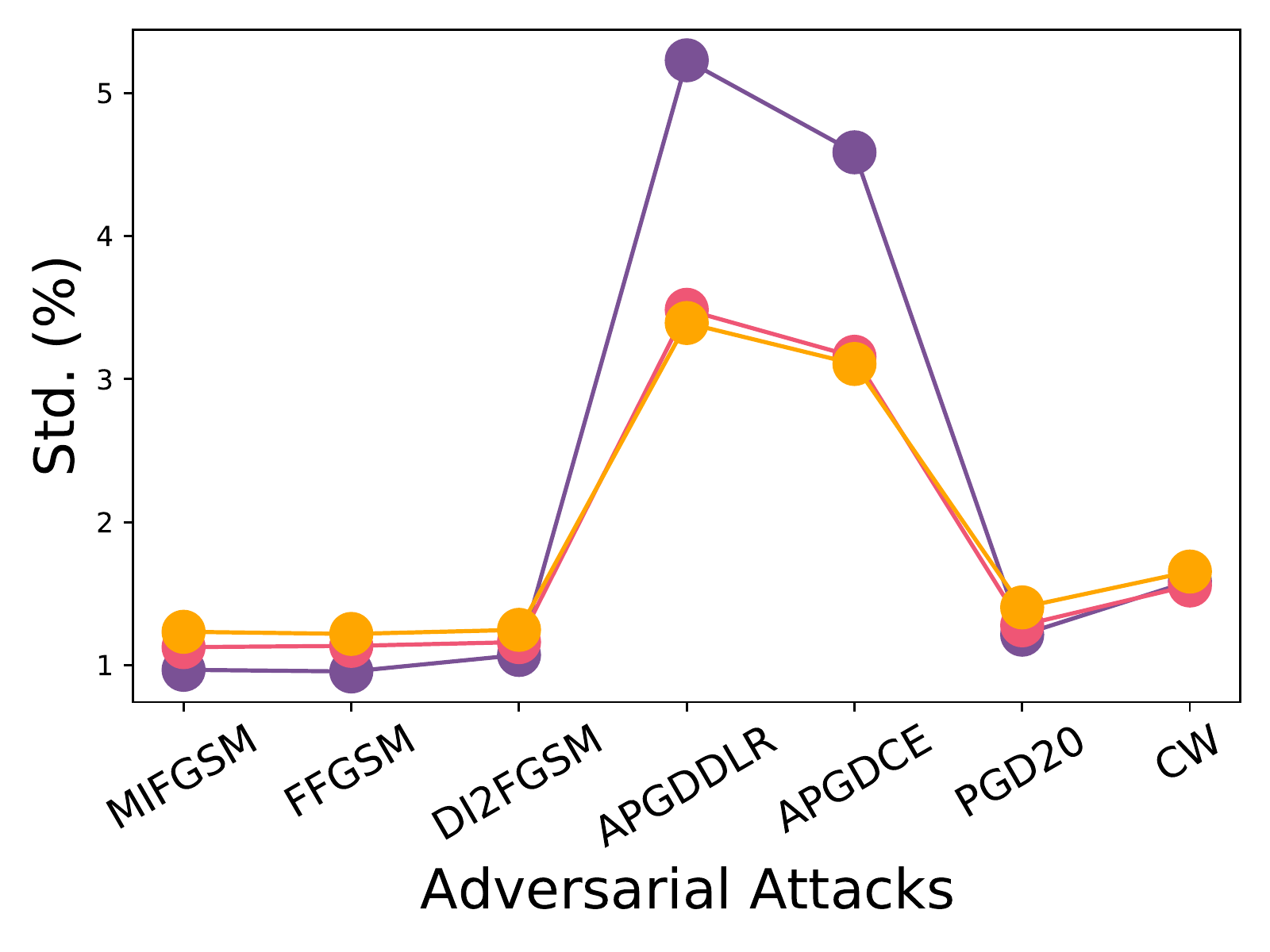} \label{fig:transfer_dev_densenet}}
\subfloat[][ResNet50]{\includegraphics[width=0.24\columnwidth]{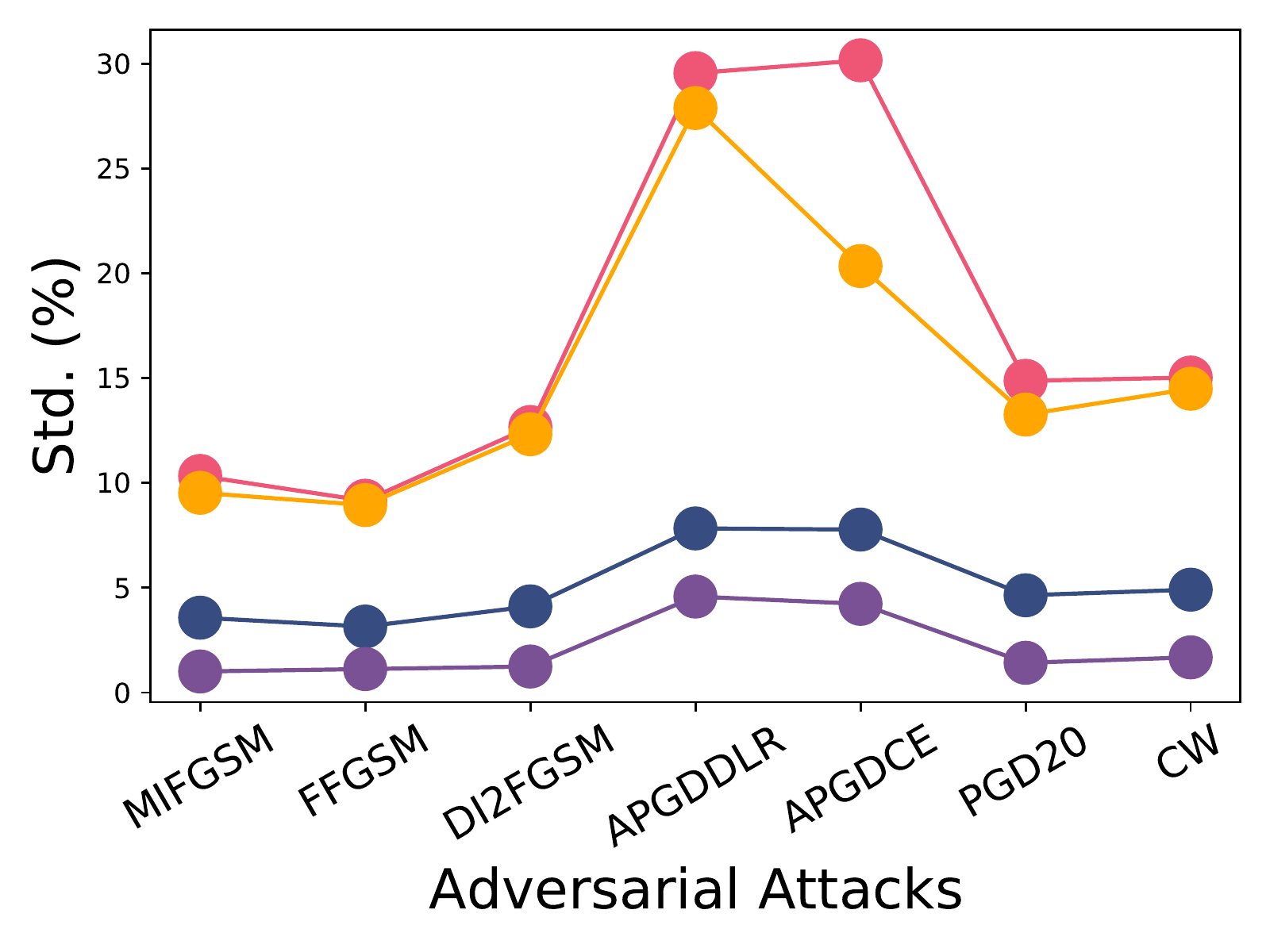} \label{fig:transfer_dev_resnet}}
\caption{Illustration of the standard deviation in performance across adversarial attacks from a variety of DNN backbones. We observe that \alg{}-based adversarial training algorithms consistently have some of the lowest deviation in performance, thus ensuring in-Transferability.}
\label{fig:std_transfer}
\end{figure}

In Figs.~\ref{fig:avg_transfer} and~\ref{fig:std_transfer}, we highlight the adversarial training algorithms with some of the lowest deviation in adversarial robustness across a number of different attacks. 
We observe that \alg{}-based adversarial training has the lowest deviation in performance for both ResNet50 and MobileNet architectures and is highly competitive on VGG16 and DenseNet.
Additionally, we observe that for APGDDLR and APDCE attacks all curves show a characteristic spike in values indicating that these attacks work well when the source and target DNNs have the same architecture but fail on dissimilar models.
Overall, we see that the \alg{}-based adversarial training algorithms consistently have low deviation in adversarial accuracy.
This trend matches the improvement in average adversarial accuracy as well, highlighting \alg{}-based adversarial training as a definite way to ensure in-Transferability. 

\subsection{Time Efficiency Comparison}
\label{subsec:efficiency_comparison}
\begin{table*}[ht!]
\centering
\caption{Illustration of the improvement in efficiency (minutes) offered by \alg{}. The total amount of time to iterate over miniImagenet and ILSVRC2012 is reduced when using \alg{}, regardless of the memory overhead caused by different DNNs. The addition of \alg{} atop adversarial training further highlights this phenomenon, with a significant reduction in training time}
\label{tab:efficiency}
\begin{tabular}{@{}cccccc@{}}
\toprule
\multirow{2}{*}{Dataset} &  \multirow{2}{*}{Algorithm} & \multicolumn{4}{c}{Minutes}\\
& & VGG16    & MobileNet & DenseNet  & ResNet50 \\
\midrule
\multirow{2}{*}{miniIMAGENET} & Baseline  & 7057.98 & 7245.39 &  5679.60 & 5743.79        \\
                              & \alg{}    & \textbf{6274.80} & \textbf{7172.28} &  \textbf{5655.44} & \textbf{5163.22}        \\
                              \midrule
\multirow{2}{*}{ILSVRC2012} & Baseline  & N/A & N/A &  N/A   & 9341.46        \\
                              & \alg{}    & N/A & N/A &  N/A & \textbf{9294.46}        \\
                              \midrule
\multirow{2}{*}{CIFAR-10}     & \cite{shafahi2019adversarial}   & 803.76  & 1691.05 & 2601.30 & 3778.68          \\
                              & \alg{} + \cite{shafahi2019adversarial} & \textbf{773.31} & \textbf{1554.38} & \textbf{2134.39} & \textbf{3434.45}         \\ 
                              \bottomrule
\end{tabular}
\end{table*}

The third of our targets deals with efficiency, specifically decreasing the amount of time taken by the training phase.
 To understand the impact of \alg{} on efficiency, we observe the time taken to iterate over 100 or more epochs across miniImagenet, ILSVRC2012 and CIFAR-10.
 The times were calculated by taking the average runtime over 1 epoch across 25 trials and expanding them to the total number of training epochs specific to each dataset-DNN pair.
 This process helps smooth the impact of variable execution loads in the computational node used to evaluate run-times.
We focus on the time taken to load miniImagenet and ILSVRC2012 during non-adversarial training as opposed to the other datasets since it requires processing individual image files and mimics the behavior when handling large-scale data.
Additionally, to highlight the practical advantage of \alg{} when performing adversarial training, we compare the total training time taken for the algorithm proposed in  \cite{shafahi2019adversarial}, with and without \alg{}. 

From Table~\ref{tab:efficiency} we observe that regardless of the amount of memory used to store different types of DNN, there is a significant reduction in the total time taken to iterate over miniImagenet and ILSVRC2012 when using only a subset of the training data.
The impact of this decrease in time to iterate is further highlighted when using the adversarial training algorithm proposed by Shafahi \textit{et al.}~\cite{shafahi2019adversarial}.
A reduction of up to $17.9\%$ in training time can be obtained when \alg{} is combined with ~\cite{shafahi2019adversarial}.
Overall, when combining the benefits in performance, efficiency and robustness, \alg{} manages to successfully deliver on all of our goals.

\section{Discussion}
\label{sec:discussion}

\paragraph{Efficiency Gain}
The results from Table~\ref{tab:curriculum_comparison} highlight that a subset of the training data is sufficient to outperform the mini-batch SGD training setup. 
An interesting caveat of how we generate our subset is the permanent removal of the noisy data while DIHCL performs soft sampling, where data is recycled into the training phase based on feedback from the model being trained.
This results in a slow reduction of the memory consumed as well as samples removed in DIHCL, which could limit it's impact on the efficiency gained.

\paragraph{Adversarial Response}
From Fig.~\ref{fig:at}, we observe that the adversarial response to MIFGSM and FFGSM between \alg{} and other adversarial training regimes are within $5\%$ of each other, across three of the four DNNs.
The importance of this observation is further highlighted by the smaller adversarial robustness of the ``Baseline’’ and DIHCL methods compared to all other training regimes. 
Given that we do not expose the model to any adversarial input during training, this outcome suggests a more inexpensive alternative that could complement adversarial training regimes.

\paragraph{DenseNet Performance}
Over the course of our experiments on standard Accuracy~($\%$) and adversarial robustness, DenseNet has offered the lowest improvement in performance.
We hypothesize that a stronger consideration of the effects of dense connectivity, as opposed to simple skip connections, could help further boost the improvements in performance. 
In addition, a closer inspection of DenseNet's building blocks to understand if the excessive redundant information being concatenated in successive layers is a key component of the general weakness to a variety of adversarial attacks could help us identify architecture specific traits to reduce their impact. 

\paragraph{Potential Negative Impacts}
Since we reduce the total amount of training data provided to the model, we risk losing some of the representational depth and complexity in the learned features.
This is especially important when considering the impact of weaker pretraining on downstream tasks.
In addition, our core idea revolves around removing datapoints that have a high proclivity of being ambiguous. 
The implications of the removal of such datapoints could be a reduction in the fairness of the overall model since such datapoints could be an underrepresented set of data.
From an adversarial robustness perspective, when using the $l_2$ metric distance as a sensitivity measure, we risk exposing our feature embeddings to alternative forms of adversarial attack.



\section{Conclusion}
\label{sec:conclusion}

Overall, we establish \alg{} as an algorithm that simultaneously tackles improvements in performance, efficiency and adversarial robustness.
The use of noise-injection in \alg{} to identify and remove noisy samples helps modify the feature embedding learned by DNNs in a favorable manner. 
In doing so, there is a strong improvement in classification accuracy achieved via a more efficient training process.
We also establish high adversarial robustness and in-Transferability by incorporating \alg{} like a plug-and-play module atop existing adversarial training methods.
An important direction of future work is exploring a variety of metrics to assess a comprehensive way to identify noisy samples.
In addition, we plan to continue pushing the capabilities of our algorithm while addressing its potential negative impacts.
Our goal is to jointly target PER in an effort to develop more cost and resource efficient training protocols, with a view to reducing the environmental impact of developing DNNs.

\section*{Acknowledgments}
This work has been partially supported (Madan Ravi Ganesh and Jason J. Corso) by NSF CI-NEW:  Collaborative Research: COVE and (Salimeh Yasaei Sekeh) by NSF CAREER 2144960, and NSF DMS 2053480; the findings are those of the authors only and do not represent any position of these funding bodies.

\bibliographystyle{IEEEbib}
\bibliography{egbib}

\appendix
\section{\alg: ILSVRC2012}
In order to efficiently execute \alg{} on ILSVRC2012, we re-purposed the algorithm to function in two phases.
In the first phase, we compute $D(.)$ (from (6) in the main manuscript)across samples of each label and reduce them using their mean value to ascertain the difference statistic over labels.
In the second phase, we refine our search space to samples across the 10 labels with the highest difference in values and capture the statistics across samples from these labels.
This is similar to assessing the prior over the 10 worst performing labels.
Doing so allows us to avoid compare statistics across a million samples, instead we simplify the comparison to samples across 10 labels, which is approximately 13000.
Thus we reduce the overall amount of memory consumed.

\section{Experimental Setup}
We provide the hyper-parameters for different baselines used in our experimental results below.

\subsection{Abbreviations}
Throughout the appendices, we use shorthand notations to simplify the discussion of certain DNN architecture or hyper-parameter names. 
We outline their full meaning below,
\begin{itemize}[topsep=0pt,itemsep=1ex,partopsep=0ex,parsep=0ex]
    \item Mob: MobileNet
    \item Dense: DenseNet
    \item R50: ResNet50
    \item Sched. : Learning rate step schedule
    \item Opt. : Optimizer
    \item Decay : Weight decay
    \item Mult. : Multiplier
    \item Mtm. : Momentum
    \item Bandit Alg. : Bandit Algorithm
    \item Loss Fb. : Loss feedback
\end{itemize}

\subsection{Curriculum Comparison}
Tables~\ref{tab:training_setup_baseline_c10_c100} and \ref{tab:training_setup_baseline_stl10_miniimagenet},  describe the hyper-parameters used for our baseline (SGD) models while Tables~\ref{tab:training_setup_dihcl_c10_c100} and \ref{tab:training_setup_dihcl_stl10_miniimagenet} describe the hyper-parameters used for the DIHCL algorithm~\cite{zhou2020curriculum}.
For the ILSVRC2012 experiments, we use Epoch=100, Batch=64, Lr=0.1, Sched. = 30,60,90, Opt.=SGD, Decay=0.00003, Mult.=0.1 and Mtm= True, with $\tau=15$.
Code for the DIHCL algorithm was provided from \url{https://github.com/tianyizhou/DIHCL}.
For \alg{}, we re-use the hyper-parameters in Tables~\ref{tab:training_setup_baseline_c10_c100} and \ref{tab:training_setup_baseline_stl10_miniimagenet} while experimenting on values for $\gamma$ and $\epsilon$, after setting $\tau=50$. The final values of $\epsilon$ and $\gamma$ for the results in Table 1 on the main manuscript are,
\begin{itemize}[topsep=0pt,itemsep=1ex,partopsep=0ex,parsep=0ex]
\item For the CIFAR-10 experiments, $\gamma=125, 125, 50, 5$ and $\epsilon=0.7, 0.1, 0.0, 0.3$ for VGG16, MobileNet, DenseNet and ResNet50 respectively.
\item For the CIFAR-100 experiments, $\gamma=250, 50, 50, 50$ and $\epsilon=0.5, 0.7, 0.1, 0.2$ for VGG16, MobileNet, DenseNet and ResNet50 respectively.
\item For the STL-10 experiments, $\gamma=12, 12, 25, 12$ and $\epsilon=0.3, 0.5, 0.0, 0.1$ for VGG16, MobileNet, DenseNet and ResNet50 respectively.
\item Finally, for the miniImagenet experiments, $\gamma=50, 125, 125, 5$ and $\epsilon=0.1, 0.7, 0.3, 0.1$ for VGG16, MobileNet, DenseNet and ResNet50 respectively.
\item Finally, for the ILSVRC2012 experiment, $\gamma=208$ and $\epsilon=0.3$.
\end{itemize}
\begin{table}[t!]
\centering
\caption{Training setups for mini-batch SGD (Baseline) on CIFAR-10 / CIFAR-100 respectively. Here, MobileNet uses cosine LR scheduling for CIFAR-100}
\label{tab:training_setup_baseline_c10_c100}
\resizebox{0.99\columnwidth}{!}{\begin{tabular}{@{}lcccc@{}}
\toprule
         & VGG16           & Mob  & Dense    & R50\\ \midrule
Epochs   & 300 / 200       & 350 / 200        & 300 / 300        & 300 / 300\\
Batch    & 128 / 128       & 128 / 128        & 64 / 64         & 128 / 128\\
Lr       & 0.1 / 0.1       & 0.1 / 0.1        & 0.1 / 0.1         & 0.1 / 0.1\\
Sched.   & 90,180,260 / 60,120,160    & 150,250 /90,180,260    & 150,225 / 150,225    & 90,180,260 /90,180,260\\
Opt.     & SGD / SGD            & SGD / SGD        & SGD / SGD         & SGD / SGD\\
Decay    & 0.0005 / 0.0005          & 0.00004 / 0.0001    & 0.0001 / 0.0001      & 0.0002 / 0.0002\\
Mult.    & 0.2 / 0.2            & 0.1 / 0.2       & 0.1 / 0.1         & 0.1 / 0.1\\
Mtm.     & True / True            & False / True      & False / False       & True / True\\
\bottomrule
\end{tabular}}
\end{table}

\begin{table}[t!]
\centering
\caption{Training setups for mini-batch SGD (Baseline) on STL-10 / miniImagenet respectively}
\label{tab:training_setup_baseline_stl10_miniimagenet}
\resizebox{0.99\columnwidth}{!}{\begin{tabular}{@{}lcccc@{}}
\toprule
         & VGG16           & Mob  & Dense    & R50\\ \midrule
Epochs   & 300 / 300       & 450 / 200        & 450 / 300        & 1000 / 300\\
Batch    & 32 / 64       & 64 / 128        & 64 / 64         & 128 / 128\\
Lr       & 0.01 / 0.01       & 0.1 / 0.1        & 0.1 / 0.1         & 0.1 / 0.1\\
Sched.   & 200 / 90,180,260    & 300,400 /90,180,260    & 300,400 / 150,225    & 300,400,600,800 /90,180,260\\
Opt.     & SGD / SGD            & SGD / SGD        & SGD / SGD         & SGD / SGD\\
Decay    & 0.0005 / 0.0005          & 0.0005 / 0.0001    & 0.0005 / 0.0001      & 0.0005 / 0.0002\\
Mult.    & 0.1 / 0.2            & 0.2 / 0.2       & 0.2 / 0.1         & 0.2 / 0.1\\
Mtm.     & True / True            & True / True      & False / False       & True / True\\
\bottomrule
\end{tabular}}
\end{table}

\begin{table}[t!]
\centering
\caption{Training setups for DIHCL on CIFAR-10 / CIFAR-100 respectively. MobileNet uses a schedule of [0 5 10 15 20 30 40 60 90 140 210 300 350] }
\label{tab:training_setup_dihcl_c10_c100}
\begin{tabular}{@{}lcccc@{}}
\toprule
         & VGG16           & Mob  & Dense    & R50\\ \midrule
Epochs        & 300 / 300             & 350 / 300        & 300 / 300        & 300 / 300\\
Bandit Alg.   & EXP3 / EXP3            & EXP3 / EXP3       & EXP3 / EXP3        & EXP3 / EXP3\\
Mean Teacher  & True / True            & True / True       & True / True        & True / True\\
Loss Fb.      & True / True            & True / True       & True / True       & True / True \\
Batch Size    & 128 / 128             & 128 / 128        & 128 / 128         & 128 / 128\\
\bottomrule
\end{tabular}
\end{table}

\begin{table}[t!]
\centering
\caption{Training setups for DIHCL on STL-10 / miniImagenet respectively.  }
\label{tab:training_setup_dihcl_stl10_miniimagenet}
\begin{tabular}{@{}lcccc@{}}
\toprule
         & VGG16           & Mob  & Dense    & R50\\ \midrule
Epochs        & 300 / 300             & 350 / 300        & 300 / 300        & 300 / 300\\
Bandit Alg.   & UCB / TS            & UCB / TS       & UCB / TS        & UCB / TS\\
Mean Teacher  & True / True            & True / True       & True / True        & True / True\\
Loss Fb.      & False / False            & False / False       & False / False       & False / False \\
Batch Size    & 128 / 128             & 128 / 128        & 64 / 64         & 128 / 128\\
\bottomrule
\end{tabular}
\end{table}

\paragraph{Ablation: Window Functions}
In studying the effects of a variety of window functions, we observe an improvement in overall $\gamma$ as well as the final testing Accuracy~($\%$).
We list the number of filters, post sensitivity, and the $\epsilon$ used to compute the final performance for each DNN.
\begin{itemize}[topsep=0pt,itemsep=1ex,partopsep=0ex,parsep=0ex]
    \item For VGG16, we use a subset of 17 filters and $\epsilon=0.7$.
    \item For MobileNet, we use a subset of 16 filters and $\epsilon=0.5$.
    \item For DenseNet, we use a subset of 12 filters and $\epsilon=0.0$.
    \item For ResNet50, we use a subset of 12 filters and $\epsilon=0.3$. In addition, we also list a subset of the optimal results for $\tau=100$.
\end{itemize}

\subsection{Adversarial Robustness}
The adversarial training algorithms we used were cloned from \url{https://github.com/locuslab/fast_adversarial}.
Most of the adversarial attacks were cloned from \url{https://github.com/Harry24k/adversarial-attacks-pytorch} while PGD20 and CW loss-based attacks were ported from \url{https://github.com/zjfheart/Friendly-Adversarial-Training}.

\paragraph{Adversarial Attacks}
In general, we use the default settings provided for all the adversarial attacks throughout our experiments.
\begin{itemize}[topsep=0pt,itemsep=1ex,partopsep=0ex,parsep=0ex]
    \item MIFGSM: $\epsilon = 8/255.$, $\alpha = 2/255.$, decay=$1.0$, iterations=$5$.
    \item FFGSM: $\epsilon = 8/255.$, $\alpha = 10/255.$.
    \item DI2FGSM: $\epsilon = 8/255.$, $\alpha = 2/255.$, decay=$0.0$, steps=$20$, resize\_rate=$0.9$, diversity\_prob=$0.5$, random\_state=False.
    \item APGD: $\epsilon = 8/255.$, steps=$100$.
    \item CWLoss: steps=$30$, $\epsilon = 0.031$, step\_size=$0.031/4$, category='Madry', rand \_init=True.
    \item PGD20: steps=$20$, $\epsilon = 0.031$, step\_size=$0.031/4$, category='Madry',  rand \_init = True.
    \end{itemize}
    
\paragraph{Adversarial Training}
We list the hyper-parameters used to train \cite{DBLP:conf/iclr/WongRK20} and \cite{shafahi2019adversarial} in Table~\ref{tab:training_setup_fast_free_c10}.
For \alg{}, we re-use the hyper-parameters in Table~\ref{tab:training_setup_fast_free_c10}  alongside our selection of $\gamma$ and $\epsilon$ while setting $\tau=50$. Specifically,
\begin{itemize}[topsep=0pt,itemsep=1ex,partopsep=0ex,parsep=0ex]
\item For the \alg{} + \cite{DBLP:conf/iclr/WongRK20}, $\gamma=125, 125, 50, 12$ and $\epsilon=0.1, 0.7, 0.2, 0.3$ for VGG16, MobileNet, DenseNet and ResNet50 respectively.
\item For the \alg{} + \cite{shafahi2019adversarial}, $\gamma=12, 125, 5, 25$ and $\epsilon=0.5, 0.3, 0.1, 0.7$ for VGG16, MobileNet, DenseNet and ResNet50 respectively.
\end{itemize}

\begin{table}[t!]
\centering
\caption{Training setups for \cite{DBLP:conf/iclr/WongRK20}and \cite{shafahi2019adversarial} on CIFAR-10}
\label{tab:training_setup_fast_free_c10}
\begin{tabular}{@{}lcccc@{}}
\toprule
         & VGG16           & Mob  & Dense    & R50\\ \midrule
Epochs        & 300             & 350        & 300         & 300\\
Batch         & 128             & 128        & 64          & 128\\
LR min        & 0.0             & 0.0        & 0.0         & 0.0\\
LR max        & 0.1             & 0.1        & 0.1         & 0.1\\
Sched.        & Cyclic          & Cyclic     & Cyclic      & Cyclic\\
Opt.r         & SGD             & SGD        & SGD         & SGD \\
Decay         & 0.0005          & 0.00004    & 0.0001      & 0.0002\\
epsilon       & 8               & 8          & 8           & 8\\
alpha         & 10              & 10         & 10         & 10\\
delta-init    & Random          & Random     & Random     & Random\\
Mtm.          & True            & True       & True       & True\\
\bottomrule
\end{tabular}
\begin{tabular}{@{}lcccc@{}}
\toprule
          & VGG16           & Mob  & Dense    & R50\\ \midrule
Epochs    & 300             & 350        & 300         & 300\\
Batch     & 128             & 128        & 64          & 128\\
LR min    & 0.0             & 0.0        & 0.0         & 0.0\\
LR max    & 0.1             & 0.1        & 0.1         & 0.1\\
Sched.    & Cyclic          & Cyclic     & Cyclic      & Cyclic\\
Opt.r     & SGD             & SGD        & SGD         & SGD \\
Decay     & 0.0005          & 0.00004    & 0.0001      & 0.0002\\
epsilon   & 8               & 8          & 8           & 8\\
Mtm.      & True            & True       & True       & True\\
\bottomrule
\end{tabular}
\end{table}
\begin{table}[t!]
\centering
\caption{Illustration of the improvement in efficiency (iterations) offered by \alg{}}
\label{tab:efficiency_iterations}
\begin{tabular}{@{}cccccc@{}}
\toprule
\multirow{2}{*}{Dataset} &  \multirow{2}{*}{Algorithm} & \multicolumn{4}{c}{Iterations Saved}\\
& & VGG16    & MobileNet & DenseNet  & ResNet50 \\
\midrule
miniIMAGENET & \alg{}  & 195.31 & 146.48 &  488.28 & 9.76        \\
                              \midrule
ILSVRC2012 & \alg{}  & -- & -- &  --   & 276.25 \\
                              \midrule
CIFAR-10     & \alg{} + \cite{shafahi2019adversarial}   & 23.43  & 292.96 & 19.53 &     48.82      \\
                              \bottomrule
\end{tabular}
\end{table}

\subsection{Efficiency Comparison}
In Table 3 of the main manuscript, we provide comparisons between the amount of time saved by using \alg{} when compared to other standard/adversarial training approaches. 
In Table~\ref{tab:efficiency_iterations}, we provide an alternative set of comparisons based on the number of iterations reduced when using \alg. 
Here, the number of iterations reduced is computed as $\frac{\gamma \times (E-\tau)}{\text{Batch Size}}$.

\end{document}